\documentclass[journal]{IEEEtran}

\usepackage{hyperref}
\usepackage{multirow}
\usepackage{amssymb}
\usepackage{amsmath}
\usepackage{amsfonts}
\usepackage{algorithm}
\usepackage{array}
\usepackage{subfigure}
\usepackage{epsfig}
\usepackage{textcomp}
\usepackage{stfloats}
\usepackage{verbatim}
\usepackage{graphicx}
\usepackage{cite}
\hypersetup{hidelinks}

\begin{document}

\title{Temporal Consistency Learning of Inter-Frames for Video Super-Resolution}

\author{Meiqin Liu, Shuo Jin, Chao Yao, Chunyu Lin,~\IEEEmembership{Member,~IEEE} and Yao Zhao, ~\IEEEmembership{Senior Member,~IEEE}
\thanks{This work was supported in part by the National Natural Science Foundation of China under Grant 61972028, Grant 61902022, and Grant 62120106009.\emph{(Corresponding author: Chao Yao.)}

Meiqin Liu, Shuo Jin, Chunyu Lin, and Yao Zhao are with the Institute of Information Science, Beijing Jiaotong University, Beijing 100044, China, and also with the Beijing Key Laboratory of Advanced Information Science and Network Technology, Beijing 100044, China(e-mail: \url{mqliu@bjtu.edu.cn}; \url{21125180@bjtu.edu.cn}; \url{cylin@bjtu.edu.cn}; \url{yzhao@bjtu.edu.cn})

Chao Yao is with the School of Computer \& Communication Engineering, University of Science and Technology Beijing, Beijing 100083, China(e-mail: \url{yaochao@ustb.edu.cn})
}
}

\markboth{IEEE TRANSACTIONS ON CIRCUITS AND SYSTEMS FOR VIDEO TECHNOLOGY,~Vol.~XX, No.~XX, April~2022}%
{Shell \MakeLowercase{\textit{et al.}}: A Sample Article Using IEEEtran.cls for IEEE Journals}


\maketitle

\begin{abstract}
Video super-resolution (VSR) is a task that aims to reconstruct high-resolution (HR) frames from the low-resolution (LR) reference frame and multiple neighboring frames. The vital operation is to utilize the relative misaligned frames for the current frame reconstruction and preserve the consistency of the results. Existing methods generally explore information propagation and frame alignment to improve the performance of VSR. However, few studies focus on the temporal consistency of inter-frames. In this paper, we propose a Temporal Consistency learning Network (TCNet) for VSR in an end-to-end manner, to enhance the consistency of the reconstructed videos. A spatio-temporal stability module is designed to learn the self-alignment from inter-frames. Especially, the correlative matching is employed to exploit the spatial dependency from each frame to maintain structural stability. Moreover, a self-attention mechanism is utilized to learn the temporal correspondence to implement an adaptive warping operation for temporal consistency among multi-frames. Besides, a hybrid recurrent architecture is designed to leverage short-term and long-term information. We further present a progressive fusion module to perform a multistage fusion of spatio-temporal features. And the final reconstructed frames are refined by these fused features. Objective and subjective results of various experiments demonstrate that TCNet has superior performance on different benchmark datasets, compared to several state-of-the-art methods.
\end{abstract}


\section{Introduction}
\IEEEPARstart{V}{ideo} Super-Resolution (VSR) is a challenging task which tries to learn the complementary information across video frames. Compared with Single Image Super-Resolution (SISR), VSR has to deal with a sequence, made up by temporally high-related but misaligned frames. In several previous works\cite{kappeler2016video,kim2019video}, VSR was regarded as an extension of SISR where the time-series data were super-resolved by image super-resolution methods\cite{liang2021swinir} frame by frame. Obviously, the performance is always not satisfactory as the temporal information fails to be well utilized.

\begin{figure}[t]
\centering
\begin{minipage}[b]{0.45\linewidth}
  \centering
 \centerline{\epsfig{figure=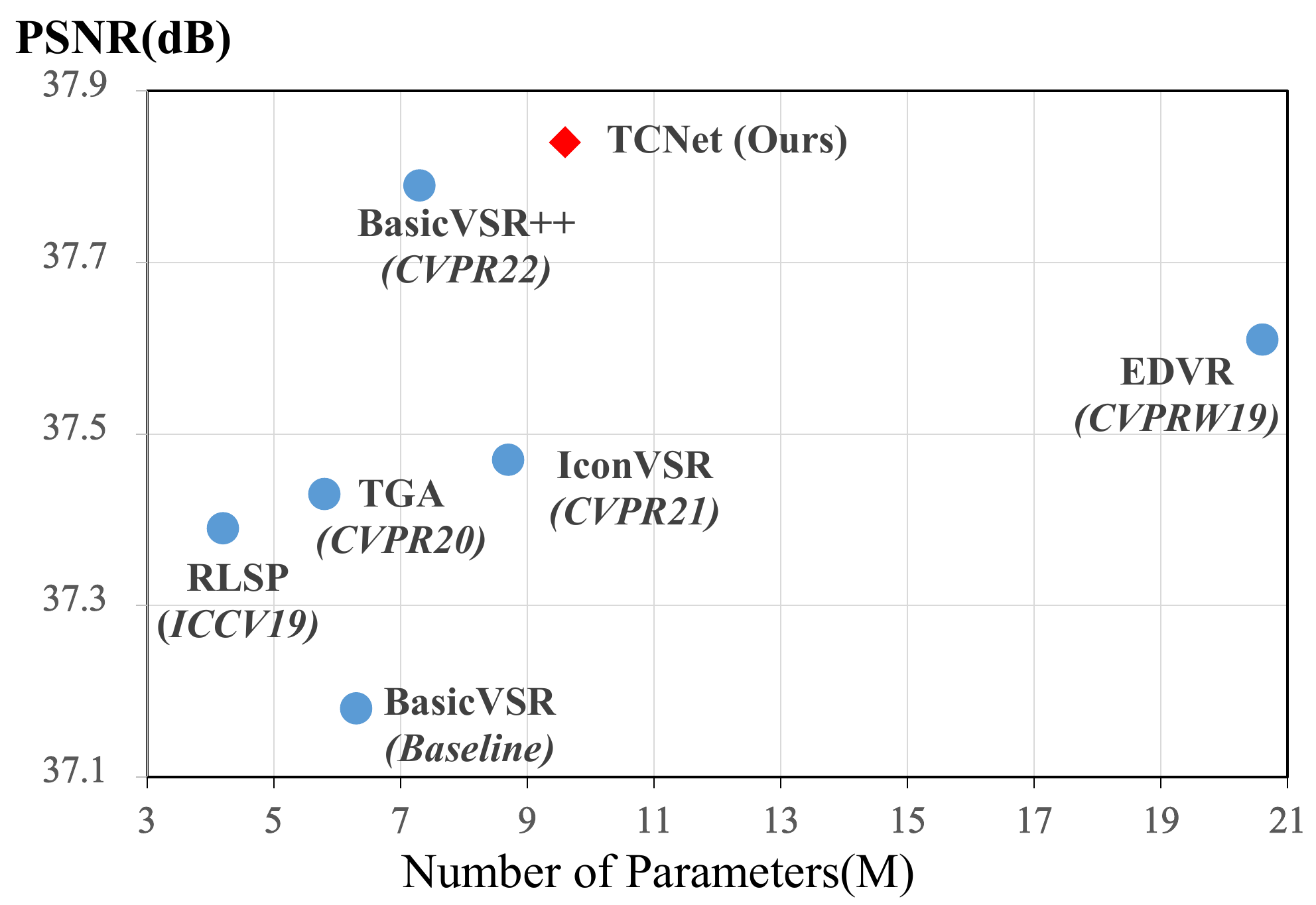,width=4.8cm}}
  \vspace{0.2cm}
  \centerline{\footnotesize (a) Degradation BI}\medskip
\end{minipage}
\hfill
\begin{minipage}[b]{0.45\linewidth}
  \centering
  \centerline{\epsfig{figure=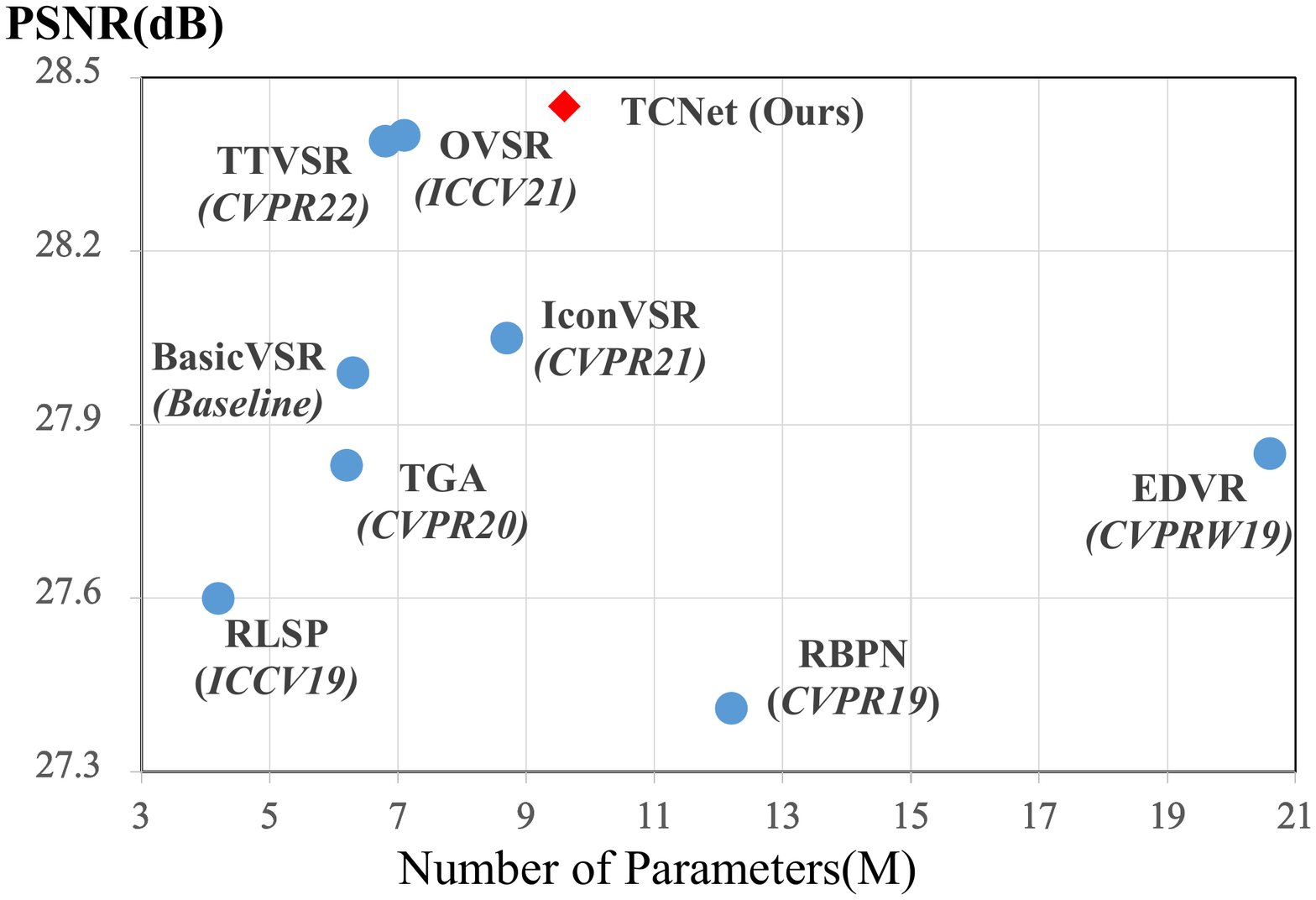,width=4.8cm}}
  \vspace{0.2cm}
  \centerline{\footnotesize (b) Degradation BD}\medskip
\end{minipage}
\caption{Quantitative performance comparison of different methods. ($a$) is the experiment with degradation BI and ($b$) is the experiment with degradation BD. More details of the experimental settings are presented in the experiment section.}
\label{fig1}
\end{figure}

Most recent VSR methods shown in Fig.~\ref{fig1} attempt to effectively leverage the temporal information from neighboring frames. One popular strategy~\cite{wang2019edvr,tian2020tdan,liu2021temporal} is the sliding-window framework, where the key frame is restored using several adjacent frames within a short temporal window. Due to the limitation of the window size, sliding window-based methods suffer from a narrow temporal scope and cannot leverage the information outside the window. 
To capture the long-term dependencies, unidirectional recurrent framework~\cite{haris2019recurrent,RSDN,sajjadi2018frame} is widely utilized as the information from distant frames can be exploited in a recurrent propagation structure. Some recent works such as BasicVSR~\cite{chan2021basicvsr} and PP-MSVSR~\cite{Jiang2021PPMSVSRMV} extend the recurrent schemes to maximize information gathering in the temporal domain. 
It is noted that the recurrent framework just leverage previous hidden states\cite{greaves2019statis}, and the extracted temporal features usually contain lots of noisy and irrelevant information, which interferes with the restoration of the current frame. Although some alignment methods have been adopted to enhance the temporal coherence, inaccurate motion estimation and compensation may still corrupt original frames and lower the performance of SR. To alleviate this problem, some instructive works~\cite{shen2021spatial,lei2020blind,zhang2020recursive} were proposed to eliminate the inconsistency between inter-frames. In particular, Shen \emph{et al.}~\cite{shen2021spatial} proposed a flow-enhanced module for temporal enhancement and a multi-frame exposure fusion technology to reduce the temporal gap. Lei \emph{et al.}~\cite{lei2020blind} proposed a novel re-weight strategy for the consistency of reconstructed videos. A convolutional network was trained with Deep Video Prior (DVP) to eliminate the flicker and offset of the video. Zhang \emph{et al.}~\cite{zhang2020recursive} proposed a novel temporal loss to ensure the consistency and a recursive block to iteratively refine features. But, the major issue of previous methods concentrates on the video coherence using the temporal prior knowledge or designing temporal loss function between frames, without considering to learn the temporal consistency features in a global range. Otherwise, the quality assessment metrics, such as PSNR and SSIM, are insensitive to overly smooth and motion blurry. The visual quality of restored videos is largely degraded by these artifacts, caused by inconsistent features.
\begin{figure}[htbp]
\centering
\begin{minipage}[b]{0.45\linewidth}
  \centering
  \centerline{\epsfig{figure=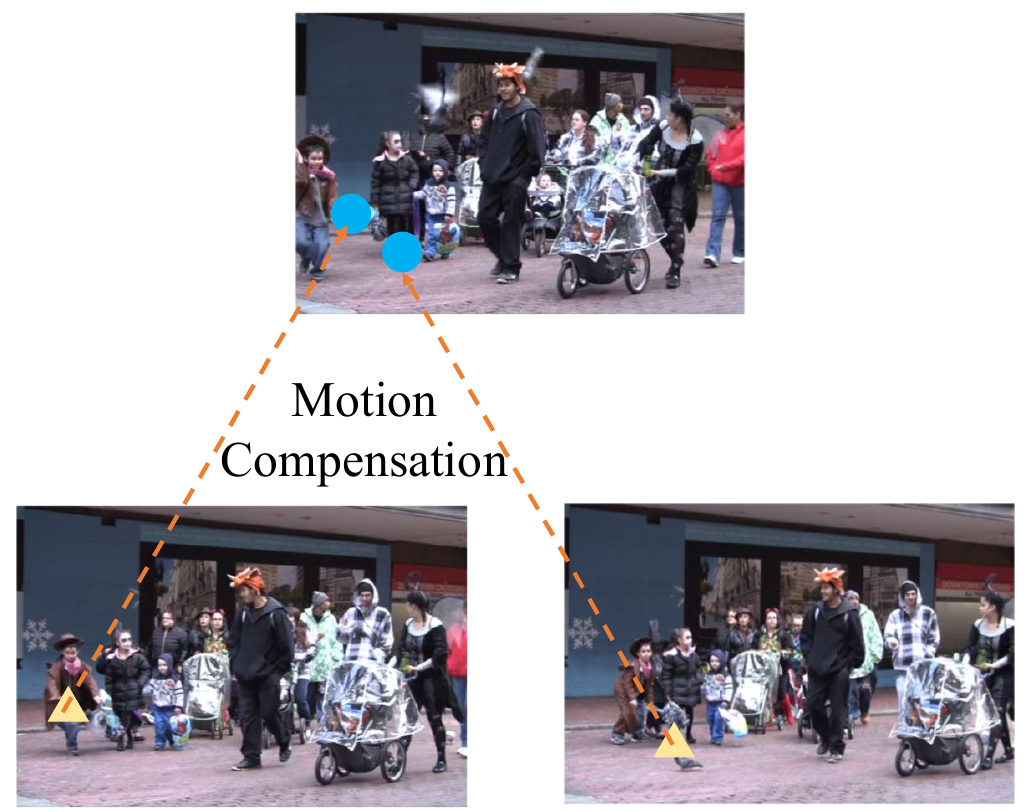,width=4.8cm}}
  \vspace{0.2cm}
  \centerline{\footnotesize (a) Local Temporal Alignment}\medskip
\end{minipage}
\hfill
\begin{minipage}[b]{0.45\linewidth}
  \centering
  \centerline{\epsfig{figure=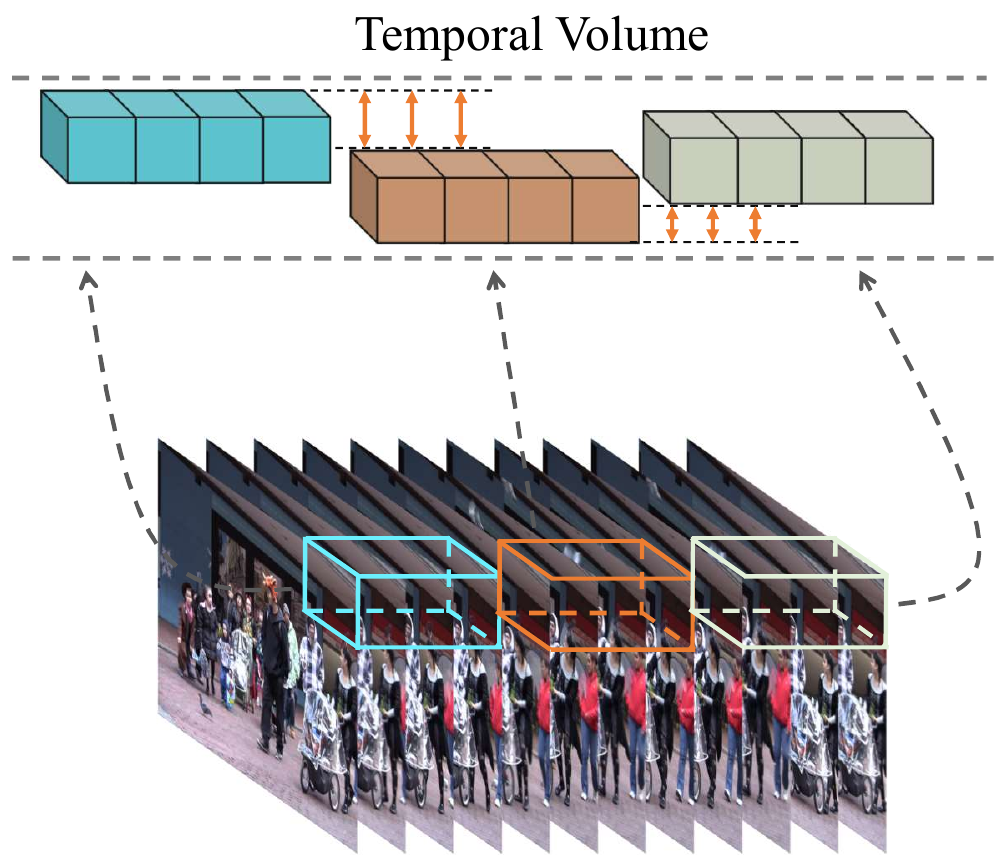,width=4.8cm}}
  \vspace{0.2cm}
  \centerline{\footnotesize (b) Global Motion Tendency}\medskip
\end{minipage}
\caption{Comparison of local temporal alignment and global motion tendency. ($a$) is the image warping operation between a reference frame and neighboring frames. ($b$) is the misalignment in the temporal volume. A temporal volume is composed of patches across multi-frames in the same location. Orange arrows denotes the displacement in the temporal volume.}
\label{intro:fig1}
\end{figure}

From the perspective of visualization, the visual content among the consecutive frames in a temporal consistent video should be stable no matter in the spatial and temporal domain. As shown in Fig.~\ref{intro:fig1}(a), the general approach is always to predict the offset between the reference frame and the previous/next frames, and then warp the adjacent frames to the current frame with optical flow or deformable convolution. In this way, the temporal inconsistency can be eliminated to some extent. Nevertheless, the temporal consistency learning is split into a number of discrete analysis frame-to-frame and limited in a local range, which is regarded as a local temporal alignment. Hence, the temporal discrepancy still exists among multiple frames.
Motivated by this, we try to learn the consistency of the feature structure in both spatial and temporal domain by considering the global motion tendency. As shown in Fig.~\ref{intro:fig1}(b), one video is split into clips across multiple frames, where each clip is located in a temporal volume. Motion tendency is the movement trajectory from a number of clips, rather than the offset between two frames. For consistency learning, stabilizing the motion tendency is a continuous analysis rather than a discrete analysis. Therefore, the temporal consistency can be learned in a global range and the temporal discrepancy of the video can be eliminated.

In this paper, we propose a temporal consistency learning approach for VSR in an end-to-end manner. A spatio-temporal stability module is designed for the continuity of global motion tendency. Specifically, the proposed module self-aligns the sequence from inter-frames, and aggregates the consistent features. In addition, a hybrid recurrent network is designed to leverage short-term and long-term information in the frame propagation. For efficiently aggregating the spatial and temporal features, a progressive fusion module is further employed to implement the feature fusion in a multi-scale space. Experimental results prove the superior performance of our proposed method on different video super-resolution benchmarks. Furthermore, we present an example of encountering short sequence for a fair comparison. The quantitative results shown in Fig.~\ref{fig1} verify that our proposed scheme improves the reconstruction quality of VSR on whether long or short sequence. The main contributions of this paper are summarized as follows:
\begin{itemize}
\item	We introduce a temporal consistency learning strategy that is designed for the alignment of temporal information from inter-frames. 
The global motion tendency is learned across multiple frames to enhance the spatio-temporal stability.
\item	We redesign a hybrid recurrent architecture for information propagation. It jointly takes advantages of sliding-window and recurrent network, which is able to exploit the short-term and long-term information.
\item	We adopt a progressive fusion module based on pyramid structure for efficiently aggregating the consistent features. The final restored frames are refined with these fused features.
\end{itemize}

\section{Related Work}
Existing VSR approaches can be mainly divided into two categories, including sliding-window framework and recurrent propagation framework.

\subsection{Sliding-window Framework for VSR}
Some early VSR  methods~\cite{narayanan2007computationally,yi2019multi,zhang2020multi,jo2018deepVS} utilized sliding-window across multiple frames for the key frame reconstruction.
In~\cite{kappeler2016video}, adjacent frames were warped to the current frame, and the convolution layers pretrained in SRCNN\cite{dong2015image} were utilized for super-resolution. Caballero \emph{et al.}~\cite{Caballero2017RealTimeVS} proposed the first VSR network in an end-to-end manner, which combined the flow estimation and SR sub-networks in training. Liu \emph{et al.}~\cite{liu2017robust} presented an alignment network to predict the spatial transformation parameters in a dynamic temporal window. TGA~\cite{TGA} divided the sequence into several groups as sliding windows, and employed 2D and 3D residual blocks for inter-window fusion. Song \emph{et al.}~\cite{song2021multi} proposed a multi-stage fusion network to fuse the temporally aligned features. Besides motion estimation and compensation (MEMC), some researchers explored non-local based methods to exploit the temporal correspondence. A non-local operation was introduced in~\cite{yi2019progressive} to capture the inter-frame information without MEMC. Zhou \emph{et al.}~\cite{zhou2021video} followed to present another non-local block for spatio-temporal information alignment. Cao \emph{et al.}~\cite{cao2021video} firstly utilized Transformer in VSR with bidirectional propagation in the feed forward network. As an alternative to optical flow, several methods employed the deformable convolution (DCN)\cite{dai2017deformable,chan2020understanding} for frame alignment. Tian \emph{et al.}~\cite{tian2020tdan} firstly adopted DCN within the sliding window for temporal alignment. Wang \emph{et al.}~\cite{wang2019edvr} proposed a multi-scale deformable convolution for a enhanced alignment and utilized a pyramid structure to improve the ability of feature fusion. Lin \emph{et al.}~\cite{lin2021fdan} further proposed a DCN module guided by the optical flow, to enhance the temporal alignment in the feature space. However, sliding-window only contains short-term information and cannot utilize more frames for reconstruction, where the frames outside the temporal window are omitted.

\subsection{Recurrent propagation framework for VSR} Since each video contains a large number of frames, it is considerate to utilize the long-term information from distant frames for the current frame reconstruction. Recently, some methods~\cite{huang2015bidirectional,isobe2020revisiting,tao2017detail} employed a recurrent framework to process the video sequence frame by frame, where the hidden states were used to transmit long-term information. Tao \emph{et al.}~\cite{tao2017detail} proposed a detail-revealing video super-resolution network (DRVSR), where the inter-frame temporal information was reserved by a convolutional long short-term memory (ConvLSTM) module~\cite{shi2015convolutional}. Sajiadi \emph{et al.}~\cite{sajjadi2018frame} designed a fast recurrent network with upsampled optical flow, which utilized the restored frames for the current frame super-resolution. Yan \emph{et al.}~\cite{yan2019frame} blended the temporal window and unidirectional framework to present a context network for the previous SR information and local information incorporating. Li \emph{et al.}~\cite{fuoli2019efficient} proposed a Recurrent Latent Space Propagation (RLSP), which concatenated the input frames and hidden states in channel dimension and transmitted historical information in feature space. RSDN~\cite{RSDN} adopted a recurrent structure-detail block to learn structure and detail information of each frame. Nevertheless, only the previous information is leveraged in the early unidirectional recurrent architecture. Some researchers start to explore the new recurrent structures to well utilize the temporal information. Huang \emph{et al.}~\cite{huang2015bidirectional} designed a bidirectional architecture to exploit the inter-frame correspondence without MEMC for alignment. BasicVSR~\cite{chan2021basicvsr} introduced a concise bidirectional recurrent network with several effective components for information refill and aggregation. BasicVSR++~\cite{chan2021basicvsr++} further presented an developed structure with enhanced propagation and alignment. Yi \emph{et al.}~\cite{yi2021omniscient} proposed an omniscient framework that employ the supporting frames from the past and future for auxiliary reconstruction. Moreover, Li \emph{et al.}~\cite{fuoli2022fast} proposed a real-time VSR network based on a recurrent attention pyramid, where the cross-attention is utilized for the temporal alignment from inter-frames. Liu \emph{et al.}~\cite{Liu2022Trajectory} further proposed a novel Trajectory-aware Transformer that is employed in a recurrent structure, where the most relevant visual tokens that located in the same trajectory are followed for long-range video modeling. Xiao \emph{et al.}~\cite{xiao2021space} proposed a space-time distillation method for performance enhancement, where the ability of large models to utilize spatio-temporal information can be transferred to light models. Although recurrent framework is able to obtain long-term information on the sequence, it only utilizes the contiguous frames, with less short-term information compared with sliding-window.    

\begin{figure*}[ht]
\begin{center}
\includegraphics[width=1.0\textwidth]{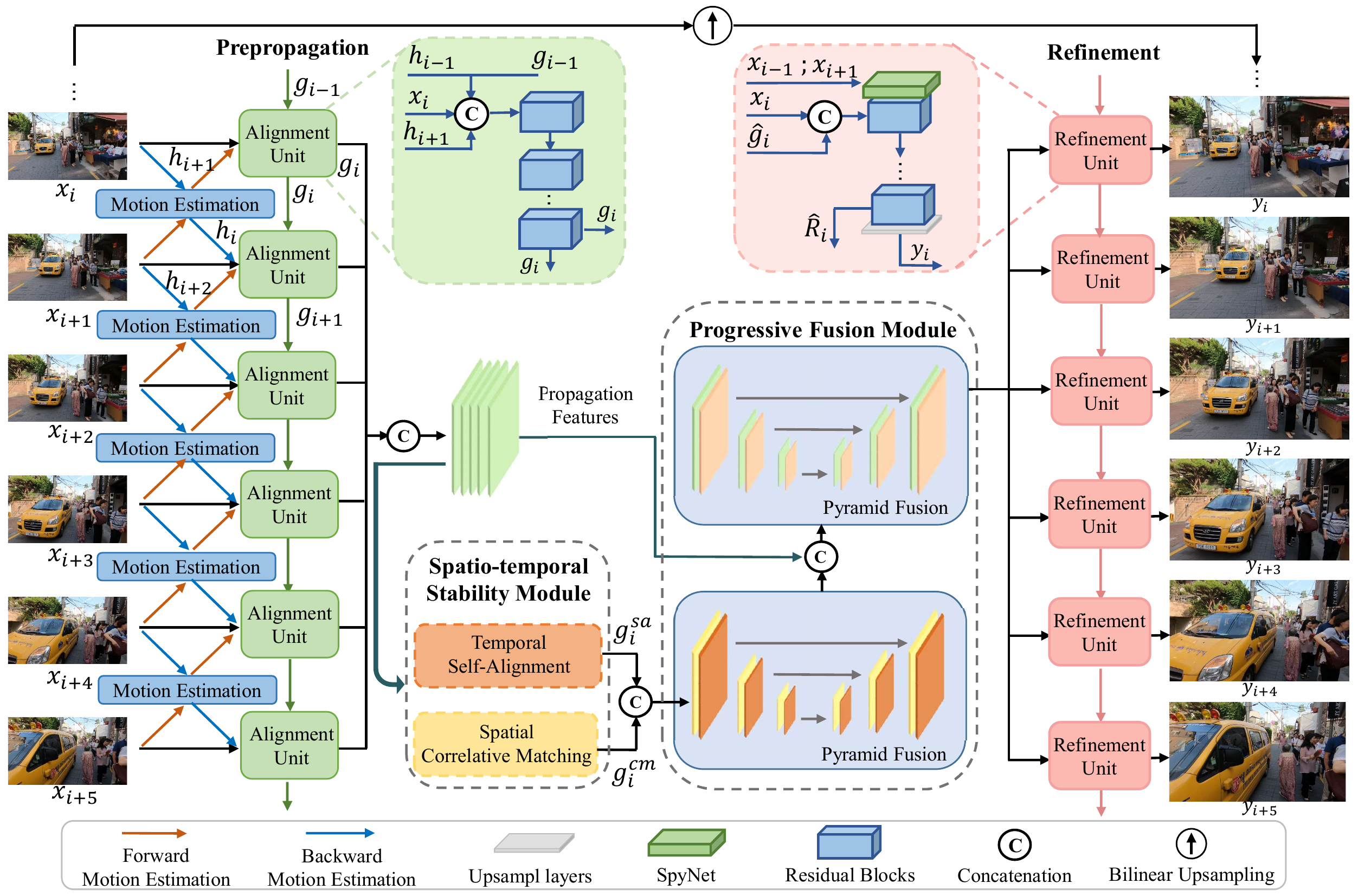}
\caption{Overview of the proposed TCNet, which is mainly made up with information pre-popagation, spatio-temporal stability module, progressive fusion, and refinement.\label{fig3}}
\end{center}
\end{figure*}
\section{Proposed Methods}
\subsection{Overview}
The overall pipeline of our proposed network is shown in Fig.~\ref{fig3}. For a given video sequence, we first adopt a hybrid recurrent framework for information prepropagation, where a pre-trained SpyNet~\cite{ranjan2017optical} is utilized for bidirectional alignment in each alignment unit. In particular, $x_i$ is taken as the input frame, $h_{i-1}$ and $h_{i+1}$ are the aligned features warped from bidirectional adjacent frames $x_{i-1}$ and $x_{i+1}$ respectively. $g_i$ is the corresponding feature extracted by several resblocks. The propagation feature is generated as follows:
\begin{equation}
    g_i = \text{Res}(cat(h_{i-1}, x_i, g_{i-1}, h_{i+1})),
\end{equation}
where $cat$ denotes the concatenation operation and $\text{Res}$ denotes the resblocks. More details about the motion estimation and compensation are introduced in the following subsection. 

Then, the propagation features are concatenated to form some groups of multiple frames, where frames in a group share the same motion tendency. A spatio-temporal stability module is employed among these groups to learn the temporal consistency in both spatial and temporal domain. In the proposed module, a spatial correlative matching mechanism is designed for structure stability. In the temporal domain, self-attention is adopted to learn the self-alignment among groups. The spatio-temporal stability module can be formulated as:
\begin{equation}
\begin{split}
    g_i^{cm} &= S_{cm}(g_i) \\
    g_i^{sa} &= I(T_{sa}(E(g)))
\end{split}
\end{equation}
where $g_i^{cm}$ and $g_i^{sa}$ denote the spatial matching and temporal aligned features respectively, $g$ denotes the concatenated features, $S_{cm}$ and $T_{sa}$ denote the spatial correlative matching and the temporal self-attention operation respectively, and $E$ denotes the embedding operation. More implementation details are presented in the following subsection.

Next, the learned stable features are progressively fused in the multi-scale space, which can be represented as:
\begin{equation}
    \hat{g_i} = \text{PF}(g_i^{cm}, g_i^{sa}, g_i),
\end{equation}
where $\hat{g_i}$ denotes the $i$-th fusion feature map, $\text{PF}$ denotes the progressive fusion. In particular, as shown in Fig.~\ref{fig3}, spatial and temporal features $g_i^{cm}$ and $g_i^{sa}$ are fused as the first stage fusion. The obtained feature $g_i^{mid}$ is then fused with propagation feature $g_i$ with the same structure, namely the second stage fusion. For clarity, we take the first stage fusion for example. First, $g_i^{cm}$ and $g_i^{sa}$ are concatenated to obtained $g_i^{c}$. Then, a resblock is used for feature extraction and a down-sampling layer is applied to obtain the LR feature, where the spatial resolution is reduced by a $4 \times 4$ convolution with stride $2$. Next, we obtain the feature with the same size by cascaded resblocks with a upsampling layer, where a $2 \times 2$ transposed convolution with stride $2$ is used to enlarge the spatial resolution. The pyramid structure fusion process can be formulated as:
\begin{equation}
\begin{split}
    g_i^c &= cat(g_i^{cm}, g_i^{sa}) \\
    g_i^{mid} &= \text{H}^{\uparrow}(\text{Res}(\text{H}^{\downarrow}(\text{Res}(g_i^c))))+g_i^c,
\end{split}
\end{equation}
where $g_i^c$ denotes the concatenated feature from the spatio-temporal stability module, $\text{H}(\cdot)^{\downarrow}$ and $\text{H}(\cdot)^{\uparrow}$ denote the down-sampling layer and upsampling layer.

Finally, another hybrid recurrent architecture is utilized for the temporally consistent features refinement. In this way, a large deal of temporally distributed information is fused with the input LR frames in the refinement units. The refinement operation can be formulated as follows:
\begin{equation}
    \hat{R_i} = \text{Res}(cat(h_{i-1}, x_i, h_{i+1}, \hat{g_i})),
\end{equation}
where $\hat{R_i}$ denotes the reconstructed feature for upsampling, $h_{i-1}$ and $h_{i+1}$ denote the warped frames with realignment. Reconstructed features are upsampled with pixel shuffle~\cite{shi2016real} to generate the high-resolution frames.

\subsection{Hybrid Recurrent Architecture}
\begin{figure*}[htbp]
\begin{center}
\includegraphics[width=0.85\textwidth]{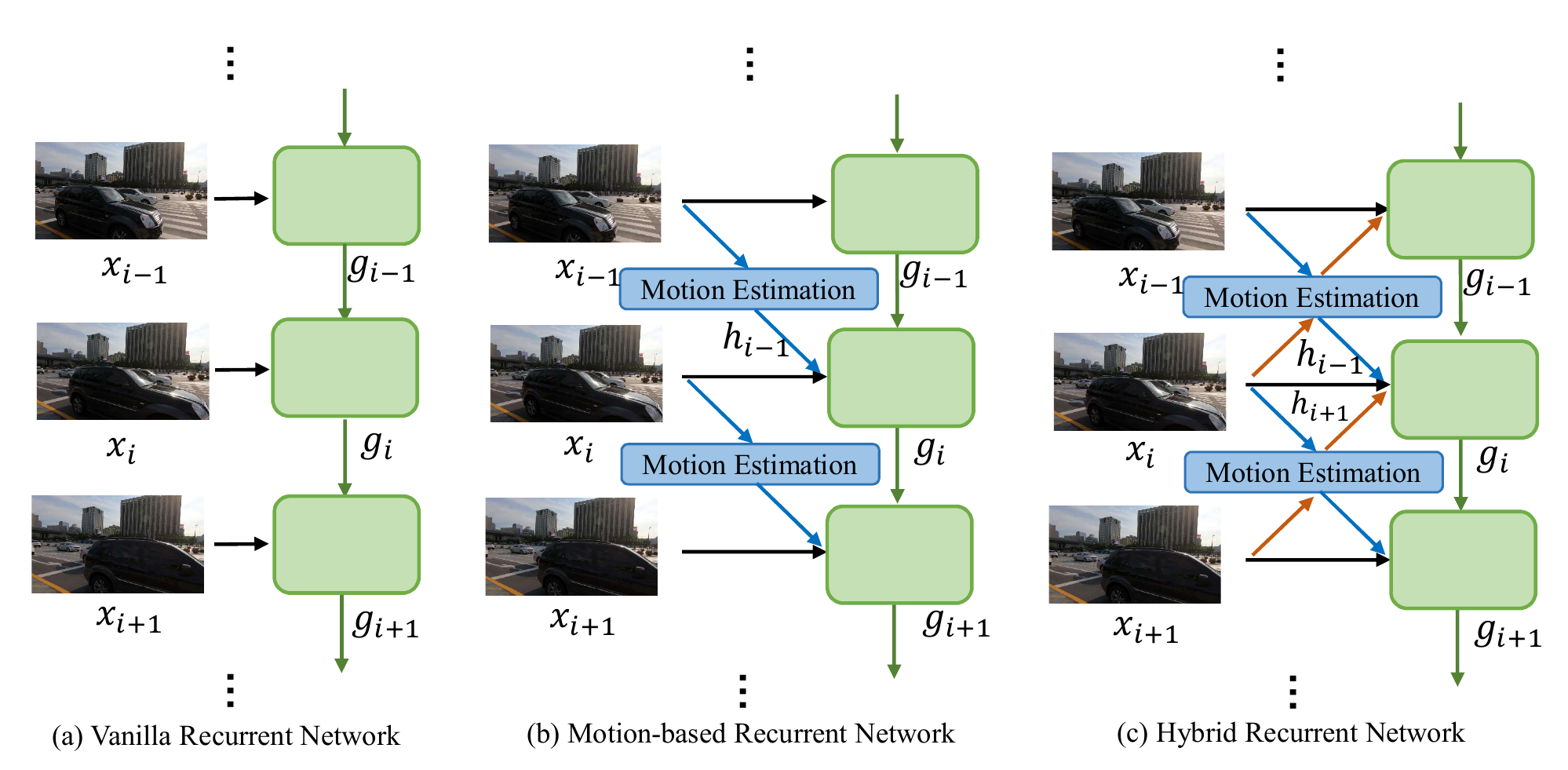}
\caption{Variants design of recurrent network. Color arrows represent the flow estimation between neighboring frames. Blue arrows represent the forward flow estimation and orange arrows represent the backward flow estimation. \label{fig2}}
\end{center}
\end{figure*}

Recurrent network has been widely used due to the effectiveness on the information propagation. However, there are still some challenges for the recurrent architecture, such as position-shifting and information oblivion. As shown in Fig.~\ref{fig2}(a), each unit in the recurrent network only employs the previous hidden states without alignment frame by frame. Thus, temporal information of inter-frames in the propagation is misaligned in the vanilla recurrent network, which may result in improper visual quality. To obtain aligned frames as additional references, the motion estimation module is available to generate aligned references, like the motion-based recurrent network shown in Fig.~\ref{fig2}(b). In this way, previous features are warped to the current feature by motion estimation. Offsets between neighboring frames can be reduced by the temporal alignment. 

Nevertheless, both two recurrent architectures only leverage the previous information and fail to utilize subsequent frames for auxiliary reconstruction. Therefore, to leverage the bidirectional information, we further propose a hybrid recurrent network, which combines the advantages of sliding-window and recurrent architecture. As shown in Fig.~\ref{fig2}(c), the hybrid recurrent architecture employs the current frame and bidirectional adjacent frames with bidirectional motion estimation, which forms a temporal window for short-range information. Specifically, the bidirectional motion estimation contains a forward and a backward motion estimation. For an input video, $x_i$ is taken as the input frame, and the forward aligned features $h_{i-1}$ is formulated as:
\begin{equation}
    h_{i-1} = \text{W}(\text{M}(x_{i-1}, x_i), x_{i-1}),\label{eq6}
\end{equation}
where $\text{M}$ and $\text{W}$ denote the motion estimation and warping operation. Eq.~\ref{eq6} also represents the alignment operation between two frames in the motion-based recurrent network. Besides, the backward aligned features $h_{i+1}$ can be generated with a similar formulation:
\begin{equation}
    h_{i+1} = \text{W}(\text{M}(x_{i+1}, x_i), x_{i+1}),\label{eq7}
\end{equation}
Based on Eq.~\ref{eq6} and Eq.~\ref{eq7}, the alignment between adjacent frames can be implemented by bidirectional motion estimation, and the adjacent frames are warped to the current frame, where a temporal sliding-window is formed. Hence, the hybrid recurrent architecture can leverage both short-term and long-term information. 

The proposed hybrid recurrent network is utilized in TCNet, where the bidirectional frame warping is utilized for a local temporal alignment. Though it is still a discrete analysis for consistency, it helps the alleviation of the consistency for global motion tendency within each video clip, compared to the other two structures. Its superior is validated in the following ablation study section. Moreover, as there are no additional neighboring frames at the beginning and end of the video sequence, we take unidirectional motion estimation for substitution and more details are presented in the experiment section. 

\subsection{Spatio-temporal Stability Module}
Video super-resolution focus on processing the video sequence to generate temporally consistent results. If the consistency is well learned, the reconstructed frames cannot be influenced by misaligned neighboring frames. Moreover, consistent reconstructed frames are smooth scrolling, continuous and share a consistent distribution in the temporal dimension. For consistent reconstruction, we propose a spatio-temporal stability module, including spatial correlative matching block (CMB) and temporal self-alignment block (TSB). 
\begin{figure}[htbp]
\begin{center}
\includegraphics[width=0.48\textwidth]{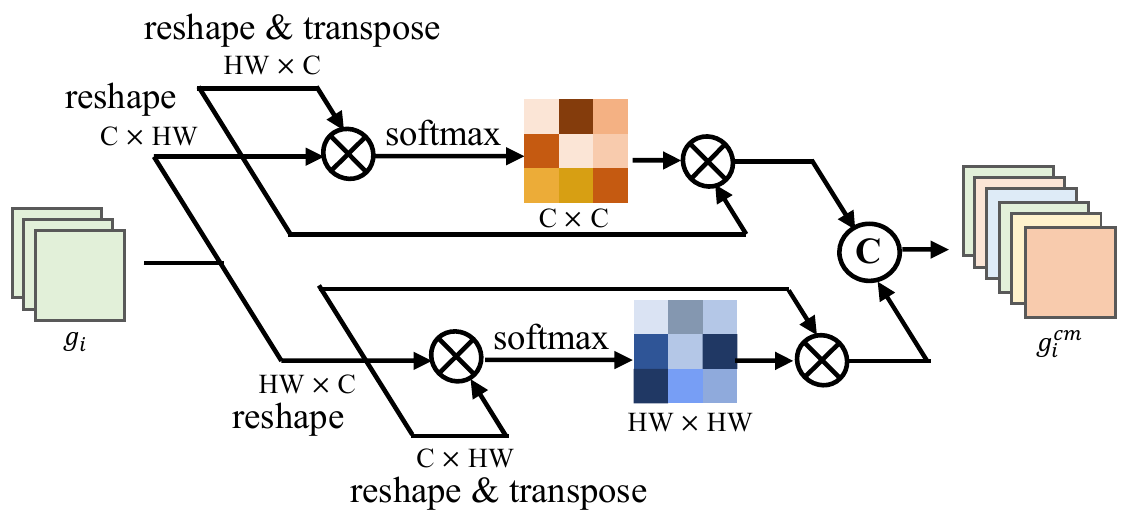}
\caption{Architecture of spatial correlative matching. Feature maps are reshaped as $C \times HW$ and $HW \times C$. They are multiplied to generate the attention matrix in parallel, where the position attention mechanism and the channel attention mechanism are implemented. \label{fignew}}
\end{center}
\end{figure}

\textbf{Spatial correlative matching.} Generally, attention mechanism attempts to obtain the similarity weights between the current region to others, by computing the weighted average of relationships from all possible regions. The similarity weights can be regarded as the matching degree. To maintain the spatial stability, we utilize a correlative matching block (CMB) to learn the long-range dependency, and implement the matching of correlative information among neighboring patches within single frame. Inspired by dual attention\cite{dualattn}, we employ both position attention mechanism (PAM) and channel attention mechanism (CAM) for a global information aggregation. The detail structure of spatial correlative matching is shown in Fig.~\ref{fignew}. In CMB, PAM and CAM are utilized in parallel for correlative matching. The obtained attention weights represent the relevance of different patches, which are used to match the correlative patches in the given frame. CMB is formulated as:
\begin{equation}
\begin{split}
        g_i^{cm} &= S_{cm}(g_i) \\
             &= \text{H}(cat(g_i, (P_{a}(g_i) + C_{a}(g_i)))),\label{eq8}
\end{split}
\end{equation}
where $g_i^{cm}$ denotes the $i$-th feature map, $\text{H}(\cdot)$ denotes the convolution layer, $cat$ denotes the concatenation operation, $P_{a}$ and $ C_{a}$ denote PAM and CAM respectively.

The matching degree among different patches is a metric for the structure stability within single frame, where the self-similarity can be measured. The fusion between the learned features from CMB and TSB is progressively implemented, where the spatio-temporal information can be aggregated. In this way, the structural stability is not only confined in the spatial domain, but also maintained in the temporal domain, guided by the matching degree of different patches across multiple frames.
\begin{figure*}[htbp]
\begin{center}
\includegraphics[width=0.92\textwidth]{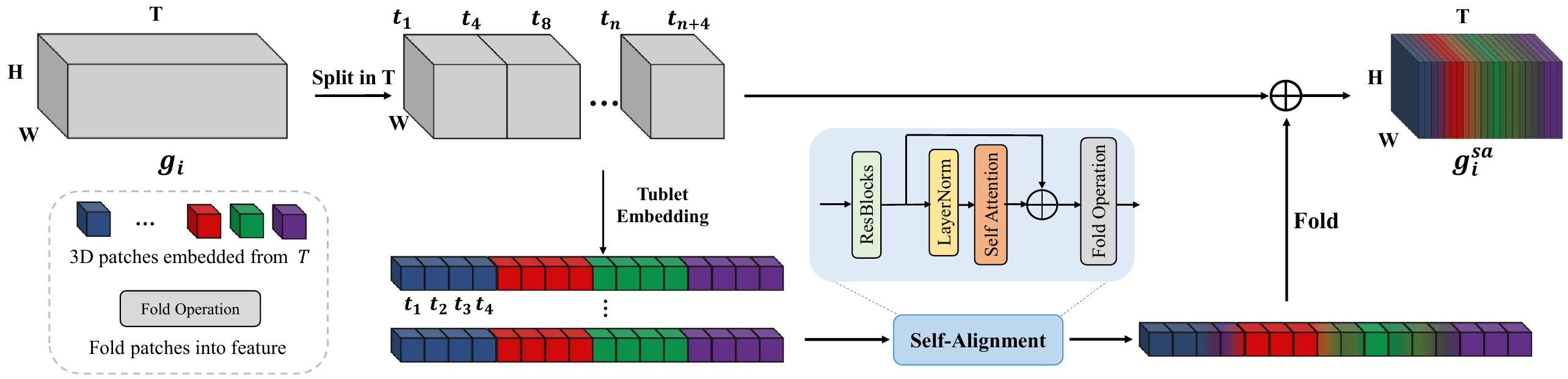}
 \caption{Architecture of temporal self-alignment. Concatenated features are firstly split in dimension $T$, and then embedded into 3D groups with height, width, and time dimension. Self-alignment is preserved by modified self-attention with fold operation (TSB). \label{fig4}}
\end{center}
\end{figure*}

\textbf{Temporal self-alignment.} Though it is simple to utilize MEMC for the alignment between two frames, it is difficult to employ MEMC to process features among multi-frames for the temporal alignment. Motivated by ViViT\cite{arnab2021vivit}, we take 3D patches among the successive frames, as shown in Fig.~\ref{fig4}. Each 3D patch contains a large deal of multi-frame correlated information. To maintain the temporal consistency, a modified self-attention block (TSB) is employed on these 3D patches to learn the self-alignment. 
\begin{figure}[htbp]
\centering
\begin{minipage}[b]{0.45\linewidth}
  \centering
  \centerline{\epsfig{figure=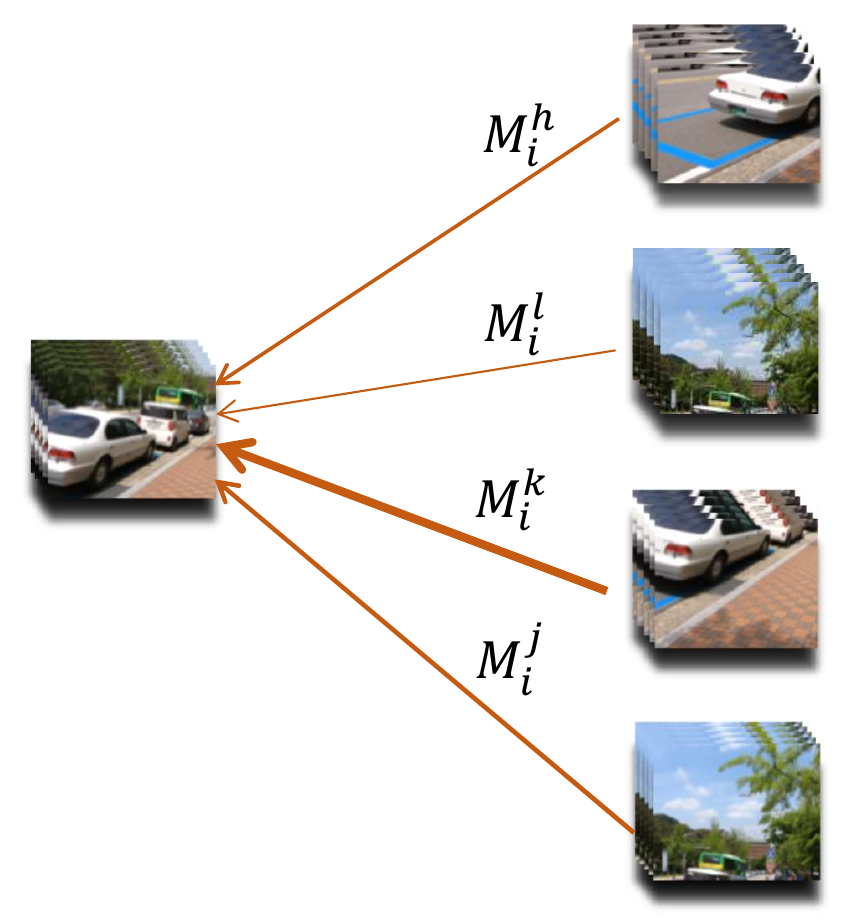,width=4.5cm}}
  \vspace{0.2cm}
  \centerline{\small (a) Self-alignment}\medskip
\end{minipage}
\hfill
\begin{minipage}[b]{0.45\linewidth}
  \centering
\centerline{\epsfig{figure=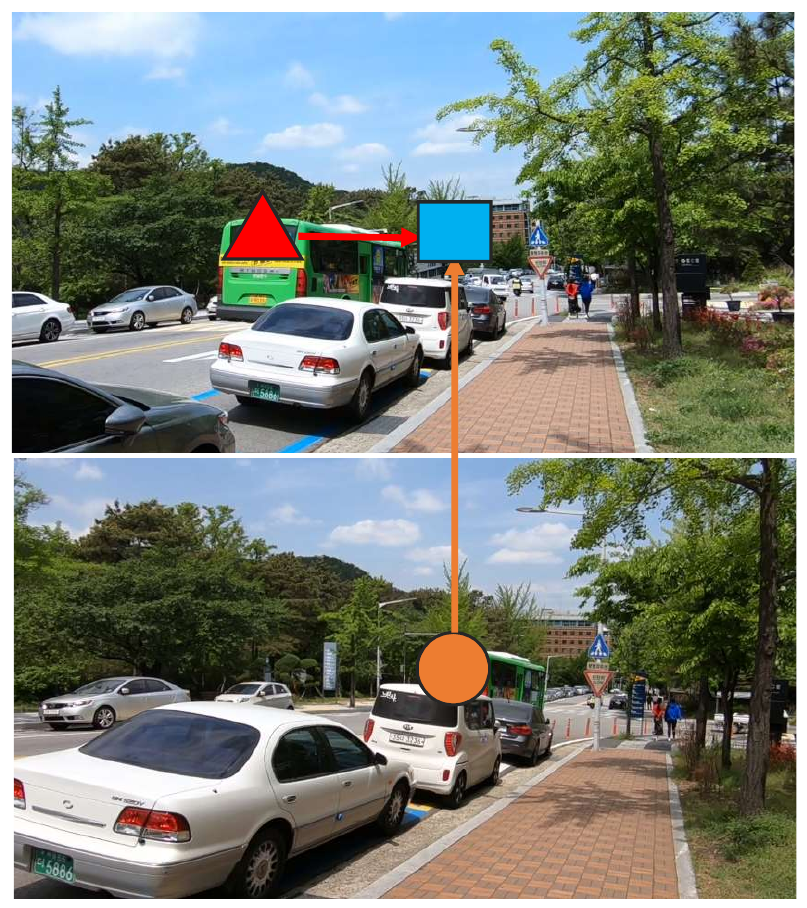,width=4.5cm}}
  \vspace{0.2cm}
  \centerline{\small (b) Image warping}\medskip
\end{minipage}
\caption{Illustration of temporal self-alignment and spatial image warping. In Fig.~\ref{fig5}(a), similarity weights between two patches are shown as solid lines, and the linewidth represents the magnitude of weights: $M_i^h$, $M_i^l$, $M_i^k$ and $M_i^j$. Fig.~\ref{fig5}(b) demonstrates the spatial warping operation between two images, where the red region is warped to the position of the blue region based on motion estimation. As $M_i^k \rightarrow 1$, and the rest $M_i^j \rightarrow 0  (j \neq k)$, the self-alignment denotes the hard warping shown in Fig.~\ref{fig5}(b), where the red region are warped to the blue region. \label{fig5}}
\end{figure}

In TSB, given the sequence feature $G\in \mathbb{R} ^{T\times C\times H\times W}$, three convolution layers are firstly employed to preprocess the input features. Patches with $t_p \times h_p \times w_p$ are extracted from the temporal, height, and width dimensions respectively, which is an extension of 2D patches to 3D patches. The number of patches is $N = n_t \times n_h \times n_w$, with $n_t = \frac{T}{t_p}$, $n_h = \frac{H}{h_p}$, $n_w = \frac{W}{w_p}$ . In this way, three 3D patch groups with $G_p\in \mathbb{R} ^{t_p \times h_p \times w_p \times CN}$ are obtained. Then, we reshape them as $G_p\in \mathbb{R} ^{N \times d}$ with $d = t_p \times h_p \times w_p \times C$ and compute the query $\mathbf{Q}$, key $\mathbf{K}$, and value $\mathbf{V}$ as:
\begin{equation}
    \mathbf{Q} = G_p W_Q, \qquad \mathbf{K} = G_p W_K, \qquad \mathbf{V} = G_p W_V, \qquad
\end{equation}
where $W_Q, W_K$ and $W_V$ denote the linear projection. $\mathbf{Q}$ is utilized to query $\mathbf{K}$ to calculate the  similarity matrix $ M \in \mathbb{R} ^{N \times N} $ as the attention map. The attention maps aggregate with the value $\textbf{V}$ to obtain the self-aligned features, which is formulated as follows:
\begin{equation}
    \text{SA}(\mathbf{Q}, \mathbf{K}, \mathbf{V}) = \text{Softmax}((\textbf{Q}^T \textbf{K})/\sqrt{d})\mathbf{V},\label{con:SA}
\end{equation}

Finally, we employ fold operation $\tau(\cdot)$ to combine these groups of 3D patches into features, with the size of $T\times C\times H\times W$ . As the obtained features are stacked in the temporal dimension $T$, index operation is utilized to separate them into independent feature maps, and the final feature map are obtained by convolution layers. To summarize, the temporal self-alignment is formulated as:
\begin{equation}
\begin{split}
        g_i^{sa} &= I(T_{sa}(E(g))) \\
             &= I(G + \tau(\text{SA}(\mathbf{Q}, \mathbf{K}, \mathbf{V}))),
\end{split}
\end{equation}
where $g_i^{sa}$ denotes the $i$-th temporal self-alignment feature, $I(\cdot)$ denotes the index operation on dimension $T$, and $\tau$ denotes the fold operation.

The relationship between different patches are presented by the similarity matrix $M$. The combination of the attention weight map and the reference feature is formulated as:
\begin{equation}
    Y_{i,j} = \sum_{i,j=1}^{N}{M_{i,j}\mathbf{V}_{i,j}},
\end{equation}
where $Y_{i,j}$ denotes the generated 3D patch, and $M_{i,j}$ denotes the $i, j$-th element of the similarity matrix. Since patches are related to dimension $T$, the feature $\textbf{V}$ contains a large number of temporal information. As shown in Fig.~\ref{fig5}(a), each 3D patch obtain a weighted response from all other patches by calculating the correlations between every two patches with self-attention mechanism. For each element of similarity matrix $M_{i,j} \in [0, 1]$, it can be treated as a soft warping operation. In the extreme case, such as $M_{i,j} = 1$, it can be regarded as a hard warping operation from one patch to the other, like the warping operation between two frames with the motion estimation. Compared with the image warping shown in Fig.~\ref{fig5}(b), which only focuses on the pixel changes of the consecutive frames, such soft warping operation focuses on the patch movement that contains the temporal information, and transmit information from other patches adaptively. In this way, obtained features among multi-frames are temporally self-aligned in the video column to maintain the consistency of the video.

\begin{table*}[htbp!]
\begin{center}
    \caption{ Quantitative comparison with BI degradation (PSNR/SSIM). Methods with * represent re-implementation. All results are calculated on Y-channel except REDS4 (RGB-channel). The best and second-best results are highlighted and underlined respectively.  Blanked entries correspond to results not reported in the papers. The runtime is computed on an LR size of 180×320.}
    \label{tab:table1}
\begin{tabular}{c|cc|ccc}
\hline
\multirow{2}{*}{Methods} & \multirow{2}{*}{Params(M)} & \multirow{2}{*}{Runtime(ms)} & \multicolumn{3}{c}{Test datasets}                                                     \\ \cline{4-6} 
                         &                                &                           & \multicolumn{1}{c}{Vid4}         & \multicolumn{1}{c}{REDS4}        & Vimeo90K-T   \\ \hline
Bicubic                  & -                              & -                         & \multicolumn{1}{c}{23.78/0.6347} & \multicolumn{1}{c}{26.14/0.7292} & 31.32/0.8684 \\
VESPCN\cite{Caballero2017RealTimeVS}                   & -                              & -                         & \multicolumn{1}{c}{25.35/0.7577} & \multicolumn{1}{c}{-}            & -            \\
TOFlow\cite{xue2019video}                   & \textbf{1.4}                            & -                     & \multicolumn{1}{c}{25.89/0.7651} & \multicolumn{1}{c}{27.98/0.7990} & 33.08/0.9054 \\
FRVSR\cite{sajjadi2018frame}                    & 5.1                            & 139                     & \multicolumn{1}{c}{26.69/0.8220} & \multicolumn{1}{c}{-}            & -            \\
DUF\cite{jo2018deepVS}                      & 5.8                            & 981                     & \multicolumn{1}{c}{27.06/0.8164} & \multicolumn{1}{c}{28.63/0.8251} & -            \\
RBPN\cite{haris2019recurrent}                     & 12.2                           & 1510                     & \multicolumn{1}{c}{27.12/0.8180} & \multicolumn{1}{c}{30.09/0.8590} & 37.07/0.9435 \\
EDVR$^*$\cite{wang2019edvr}                     & 20.6                           & 378                    & \multicolumn{1}{c}{27.35/0.8264} & \multicolumn{1}{c}{31.09/0.8800} & 37.61/0.9489 \\
PFNL$^*$\cite{yi2019progressive}                     & 3.0                            & 295                    & \multicolumn{1}{c}{26.73/0.8029} & \multicolumn{1}{c}{29.63/0.8502} & 36.14/0.9363 \\
MuCAN\cite{li2020mucan}                    & 13.6                              & 1231                         & \multicolumn{1}{c}{-}             & \multicolumn{1}{c}{30.88/0.8750} & 37.32/0.9465 \\
TGA\cite{TGA}                      & 5.8                            & 390                     & \multicolumn{1}{c}{27.19/0.8213} & \multicolumn{1}{c}{-}            & 37.43/0.9480 \\
RLSP\cite{fuoli2019efficient}                     & 4.2                            & \textbf{46}                     & \multicolumn{1}{c}{27.15/0.8202} & \multicolumn{1}{c}{-}            & 37.39/0.9470 \\
BasicVSR$^*$\cite{chan2021basicvsr}                 & 6.3                            & 63                      & \multicolumn{1}{c}{27.29/0.8267} & \multicolumn{1}{c}{31.42/0.8909} & 37.18/0.9450 \\
IconVSR$^*$\cite{chan2021basicvsr}                  & 8.7           
& 70                      & \multicolumn{1}{c}{27.39/0.8279} & \multicolumn{1}{c}{31.67/0.8948} & 37.47/0.9476 \\
VSRT\cite{cao2021video}                  & 43.8                            & 242                      & \multicolumn{1}{c}{27.36/0.8258} & \multicolumn{1}{c}{31.19/0.8815} & \underline{37.71}/\underline{0.9494} \\
TTVSR\cite{Liu2022Trajectory}                  & 6.8                            & 150                      & \underline{27.41/0.8344} & \multicolumn{1}{c}{\textbf{31.97}/\textbf{0.9007}} & 37.60/0.9502 \\
TCNet (Ours)             & 9.6                            & 94                      & \multicolumn{1}{c}{\textbf{27.48}/\textbf{0.8380}} & \multicolumn{1}{c}{\underline{31.82}/\underline{0.9002}} & \textbf{37.84}/\textbf{0.9514} \\ \hline
\end{tabular}
\end{center}
\end{table*}
\begin{table*}[htbp!]
\begin{center}
\caption{ Quantitative comparison (PSNR/SSIM) on Vid4 with BD degradation. All results are calculated on Y-channel. Methods with * represent re-implementation. The best and second-best results are highlighted and underlined respectively.}
    \label{tab:table2}
\begin{tabular}{ccccccc}
\hline
Methods     & Params(M) & Calendar(Y)  & City(Y)      & Foliage(Y)   & Walk(Y)      & Average(Y)   \\ \hline
Bicubic     & -         & 18.41/0.4619 & 23.18/0.4927 & 21.49/0.4565 & 24.12/0.6873 & 21.80/0.5246 \\
FRVSR\cite{sajjadi2018frame}       & 5.1       & 23.46/0.7854 & 27.70/0.8099 & 25.96/0.7560 & 29.69/0.8990 & 26.70/0.8126 \\
DUF\cite{jo2018deepVS}         & 5.8       & 23.85/0.8052 & 27.97/0.8253 & 26.22/0.7646 & 30.47/0.9118 & 27.13/0.8267 \\
RBPN\cite{haris2019recurrent}        & 12.2      & 24.33/0.8244 & 28.28/0.8413 & 26.46/0.7753 & 30.58/0.9130 & 27.41/0.8385 \\
TDAN\cite{tian2020tdan}        & 2.29      & 23.56/0.7896 & 27.53/0.8028 & 26.00/0.7491 & 29.99/0.9032 & 26.77/0.8112 \\
FFCVSR\cite{yan2019frame}      & -         & 24.39/0.8250 & 27.80/0.8314 & 26.70/0.7868 & 30.55/0.9124 & 27.36/0.8389 \\
EDVR$^*$\cite{wang2019edvr}        & 20.6      & 24.80/0.8368 & 28.54/0.8508 & 26.95/0.7870 & 31.12/0.9266 & 27.85/0.8503 \\
PFNL$^*$\cite{yi2019progressive}        & 3.0       & 24.37/0.8246 & 28.09/0.8385 & 26.51/0.7768 & 30.64/0.9134 & 27.41/0.8383 \\
RSDN$^*$\cite{RSDN}        & 6.2       & 24.74/0.8386 & 28.75/0.8554 & 27.00/0.8013 & 30.85/0.9183 & 27.83/0.8534 \\
TGA\cite{TGA}         & 5.8       & 24.47/0.8286 & 28.37/0.8419 & 26.59/0.7793 & 30.96/0.9181 & 27.59/0.8419 \\
RLSP\cite{fuoli2019efficient}        & 4.2       & 24.60/0.8335 & 28.14/0.8453 & 26.75/0.7925 & 30.88/0.9192 & 27.60/0.8476 \\
BasicVSR$^*$\cite{chan2021basicvsr}    & 6.3       & 24.89/0.8423 & 28.91/0.8563 & 27.17/0.7921 & 30.99/0.9326 & 27.99/0.8558 \\
IconVSR$^*$\cite{chan2021basicvsr}     & 8.7       & 24.94/0.8434 & 28.96/0.8571 & 27.22/0.7933 & 31.07/0.9334 & 28.05/0.8568 \\
PP-MSVSR\cite{Jiang2021PPMSVSRMV}    & \textbf{1.5}       & 25.03/0.8486 & 28.90/0.8524 & 27.18/0.7927 & 31.40/0.9479 & 28.13/0.8604 \\
OVSR$^*$\cite{yi2021omniscient}        & 7.1       & 25.22/0.8571 & 28.93/0.8729 & 27.43/0.8209 & 31.70/0.9301 & 28.32/0.8703 \\
TTVSR$^*$\cite{Liu2022Trajectory}        & 6.8       & \underline{25.31}/0.8537 & 28.99/0.8696 & \underline{27.53}/0.8175 & 31.76/0.9270 & 28.40/0.8695 \\
OVSR\cite{yi2021omniscient}        & 7.1       & 25.28/\underline{0.8581} & \underline{29.10}/\underline{0.8769} & 27.49/\underline{0.8230} & \underline{31.79}/\underline{0.9314} & \underline{28.41}/\underline{0.8724} \\
TCNet (Ours) & 9.6       & \textbf{25.31}/\textbf{0.8584} & \textbf{29.12}/\textbf{0.8780} & \textbf{27.53}/\textbf{0.8228} & \textbf{31.81}/\textbf{0.9326} & \textbf{28.44}/\textbf{0.8730} \\ \hline
\end{tabular}
\end{center}
\end{table*}
\section{EXPERIMENTS}
\subsection{Datasets and details}
Following previous works~\cite{chan2021basicvsr,wang2019edvr}, we adopt two widely used video datasets REDS~\cite{nah2019ntire} and Vimeo90K~\cite{xue2019video} as our training sets. For REDS, we consider REDS4 as our test set, REDSval4 as our validation set, and the remaining set as the training set. We take Vid4~\cite{vid4} and Vimeo90K-T~\cite{xue2019video} as test sets while training our network on Vimeo90K. Our networks are trained and tested with 4x down-sampling operation using degradation Bicubic (BI). We also give additional results with degradation Blur Down (BD). For BI, the MATLAB function imresize is employed for the down-sampling operation. For BD, we adopt Gaussian blur with $\sigma=1.6$ to generate the blurry ground-truth, followed by down-sampling every four pixels. Two widely used quality assessments PSNR and SSIM~\cite{wang2004image} are adopted to evaluate the performance. 
\begin{table*}[htbp]
\begin{center}
    \caption{ Quantitative comparison (PSNR/SSIM) on REDS4 for short sequence experiments. The results are trained and tested on video clips with 5 or 7 frames. The best and second-best results are highlighted and underlined respectively.}
    \label{tab:table3}
\begin{tabular}{ccccccc}
\hline
Methods       & Frames & Clip\_000    & Clip\_011    & Clip\_015    & Clip\_020    & Average      \\ \hline
Bicubic      & -            & 24.55/0.6489 & 26.06/0.7261 & 28.52/0.8034 & 25.41/0.7386 & 26.14/0.7292 \\
DUF\cite{jo2018deepVS}                 & 7      & 27.30/0.7937 & 28.38/0.8056 & 31.55/0.8846 & 27.30/0.8164 & 28.63/0.8251 \\
TOFlow\cite{xue2019video}               & 7      & 26.52/0.7540 & 27.80/0.7858 & 30.67/0.8609 & 26.92/0.7953 & 27.98/0.7990 \\
EDVR-M\cite{wang2019edvr}              & 5      & 27.75/0.8153 & 31.29/0.8732 & 33.48/0.9133 & 29.59/0.8776 & 30.53/0.8699 \\
EDVR\cite{wang2019edvr}               & 5      & 28.01/0.8250 & 32.17/0.8864 & 34.06/\textbf{0.9206} & 30.09/0.8881 & 31.09/0.8800 \\
MuCAN\cite{li2020mucan}                & 7      & 27.99/0.8219 & 31.84/0.8801 & 33.90/0.9170 & 29.78/0.8811 & 30.88/0.8750 \\
BasicVSR\cite{chan2021basicvsr}            & 5      & 27.67/0.8114 & 31.27/0.8740 & 33.58/0.9135 & 29.71/0.8803 & 30.56/0.8698 \\
IconVSR\cite{chan2021basicvsr}             & 5      & 27.83/0.8182 & 31.69/0.8798 & 33.81/0.9164 & 29.90/0.8841 & 30.81/0.8746 \\
VSRT\cite{cao2021video}             & 5      & \underline{28.06}/\underline{0.8267} & \underline{32.28}/\underline{0.8883} & \underline{34.15}/0.9199 & \textbf{30.26}/\underline{0.8912} & \underline{31.19}/\underline{0.8815} \\
TCNet(Ours)        & 5      & \textbf{28.27}/\textbf{0.8342} & \textbf{32.41}/\textbf{0.8897} & \textbf{34.29}/\underline{0.9204} & \underline{30.24}/\textbf{0.8913} & \textbf{31.30}/\textbf{0.8839} \\ \hline
\end{tabular}
\end{center}
\end{table*}

During the training stage, cosine annealing scheme is adopted, and the initial learning rate of the main network and the flow network are set to $1 \times 10^{-4}$ and $2.5 \times 10^{-5}$, respectively. The training stage stops when the loss function is no longer significantly reduced and the total number of iterations is about 400-600K. The weights of the feature extractor and flow estimator are fixed during the first 10,000 iterations. We utilize ADAM~\cite{kingma2014adam} optimizer with $\beta_1=0.9$, $\beta_2=0.999$ and $\epsilon=10^{-8}$. Patch size of each frame is set to $64 \times 64$. Batch size and num worker of each GPU is set to 4 and 6 respectively. We adopt 20 residual blocks for feature extraction. The number of CMB and TSB are both set to
8. The size of 3D patch is set to 8. The number of feature channels is set to 64. Considering the particularity of video sequences, we apply unidirectional motion estimation at the beginning and the end of sequences. For remaining frames that cannot be divisible by 3D patch size in dimension $T$, 2D patch embedding operation is adopted. Charbonnier loss~\cite{lai2017deep} is used to restrict the whole network, because of its better performance than $l1$ loss and $MSE$ loss in restoring edge details. It can be formulated as follows:
\begin{equation}
    \mathcal{L_C} = \frac{1}{N}\sum_{i = 1}^{N}\sqrt{{(I_i^{SR} - I_i^{HR})^2 + \varepsilon^2}},
\end{equation}
where $I_i^{SR}$ denotes the $i$-th reconstructed image and $\varepsilon$ is a small constant set to $10^{-3}$. All of our experiments are conducted on PyTorch framework and NVIDIA RTX A4000 GPUs. Runtime is tested on one NVIDIA RTX A4000 GPU with 16GB memory, and Hygon 2.2GHz CPU.

\begin{figure}[h]
\begin{center}
\includegraphics[width=0.45\textwidth]{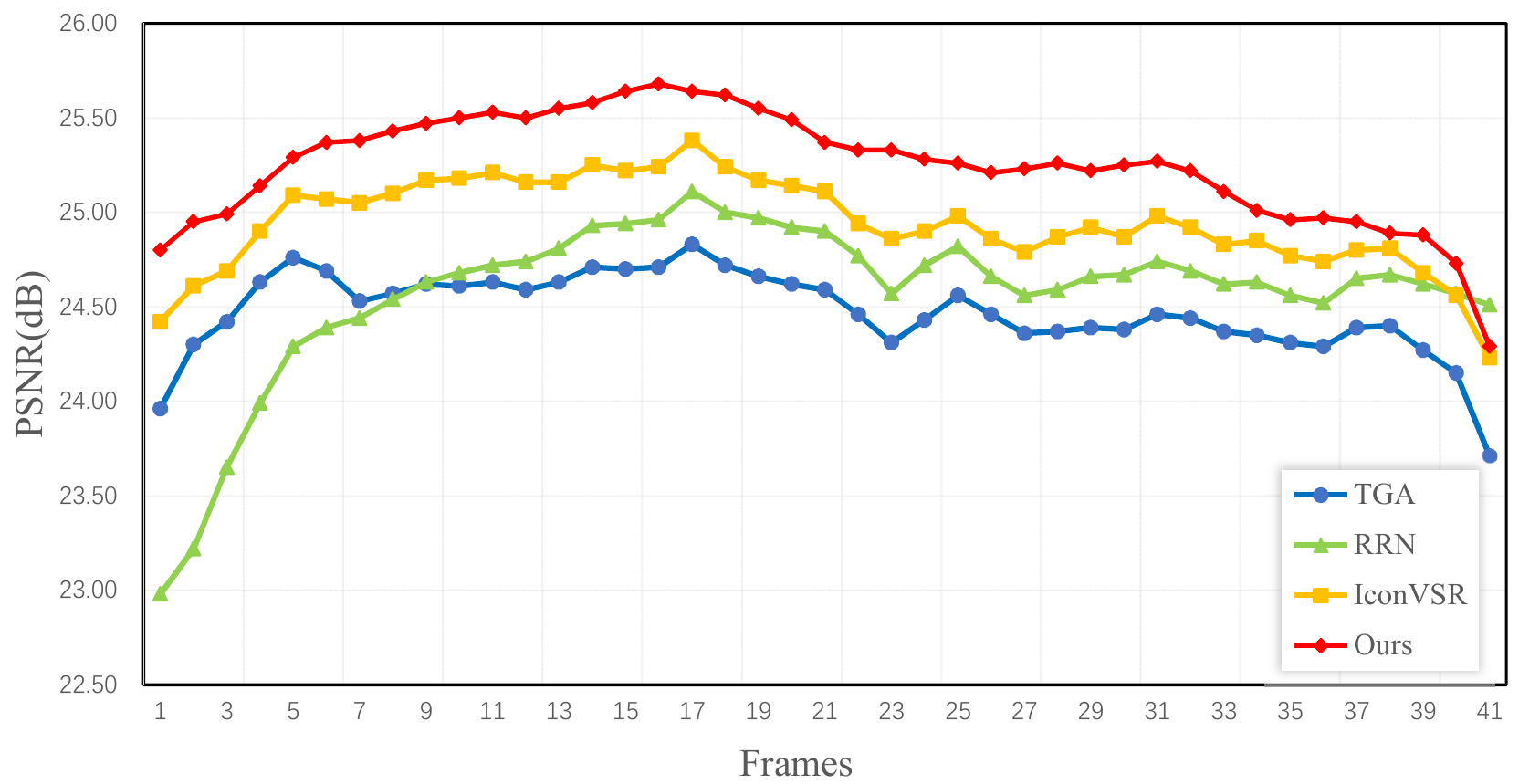}
\caption{Reconstruction results of each frame on Vid4 dataset. Blue, green and yellow lines represent the result of TGA (sliding window framework), RRN (unidirectional recurrent framework) and IconVSR (bidirectional recurrent framework) respectively.\label{fig8}}
\end{center}
\end{figure}
\subsection{Comparisons with Other Methods}
We compare TCNet with other state-of-the-art methods, such as FRVSR~\cite{sajjadi2018frame}, RBPN~\cite{haris2019recurrent}, EDVR~\cite{wang2019edvr}, PFNL~\cite{yi2019progressive}, TGA~\cite{TGA}, DUF~\cite{jo2018deepVS}, RLSP~\cite{fuoli2019efficient}, BasicVSR~\cite{chan2021basicvsr}, PP-MSVSR~\cite{Jiang2021PPMSVSRMV}, and OVSR~\cite{yi2021omniscient}. Three testing datasets contain Vid4~\cite{vid4}, Vimeo-90K-T~\cite{xue2019video} and REDS4. For a fair comparison, we unify the degradation for each experiment.

Several previous methods are re-implemented on different training datasets to obtain the corresponding results and other results are taken from their papers. As summarized in Tab.~\ref{tab:table1}, TCNet performs superior to other state-of-the-art methods on different test datasets, in terms of average PSNR and SSIM. In particular, TCNet surpasses VSRT 0.12 dB and EDVR 0.13 dB on Vid4, with fewer parameters. Besides, TCNet surpasses VSRT 0.15 dB and IconVSR 0.37 dB on Vimeo90K-T. These results demonstrate that the effectiveness of TCNet for video super-resolution. The misalignment of inter-frames can be well eliminated by the proposed hybrid recurrent propagation and spatio-temporal stability module. When compared with VSRT, TCNet outperforms it on three test datasets, with only 40\% inference time and about 78\% fewer parameters. Although TTVSR performs better than TCNet in REDS4 test set, TCNet can reduce 34\% time consumption. These results also demonstrate that our method is not relied on the stack of blocks and is efficient. Besides, TCNet gains a more promotion on SSIM than PSNR. It means the structural similarity of different frames is well maintained. It is because the proposed CMB preserves the self-similarity of the single frame with matching degree among relative patches, and inter-frame information also enhance the stability of the frames structure.
\begin{figure*}[htbp]
    \centering
    \subfigure[Subjective results on REDS4] {\includegraphics[width=0.9\textwidth]{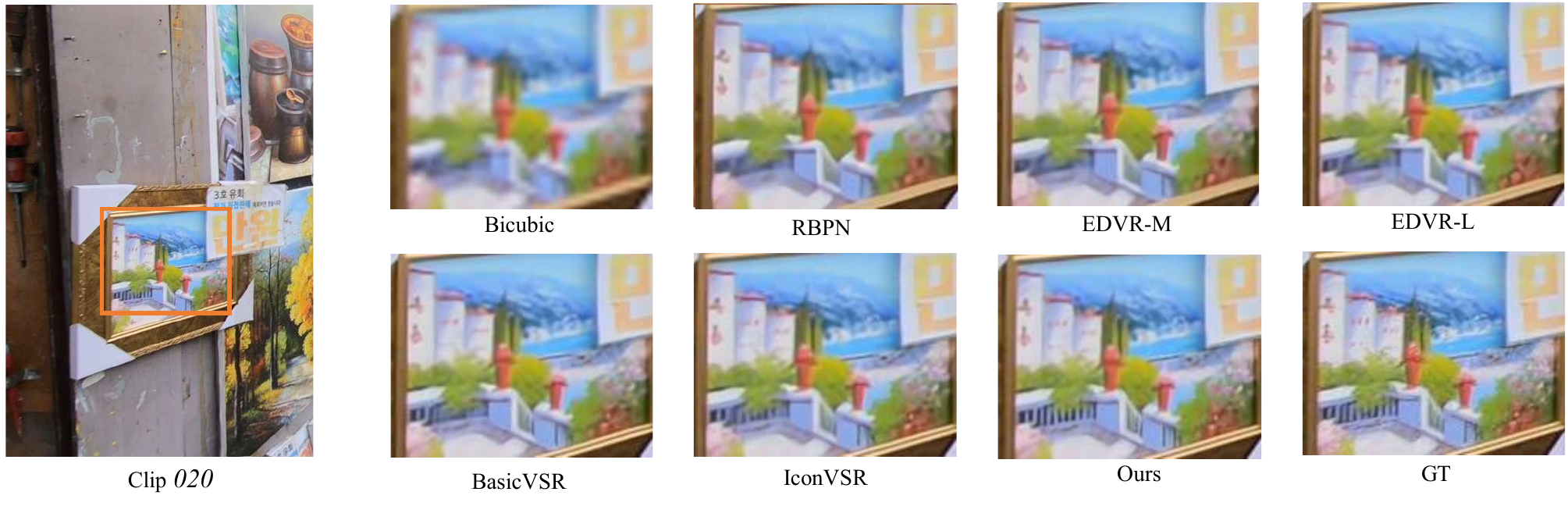}}
    \subfigure[Subjective results on Vimeo90K-T] {\includegraphics[width=0.9\textwidth]{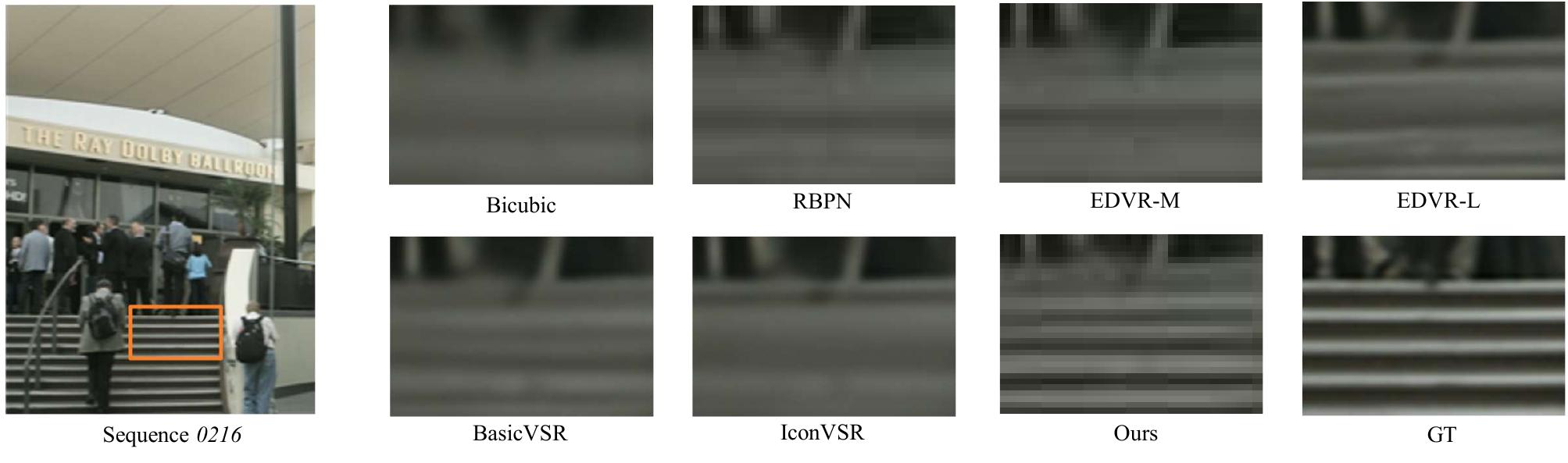}}
    \subfigure[Subjective results on Vid4] {\includegraphics[width=0.9\textwidth]{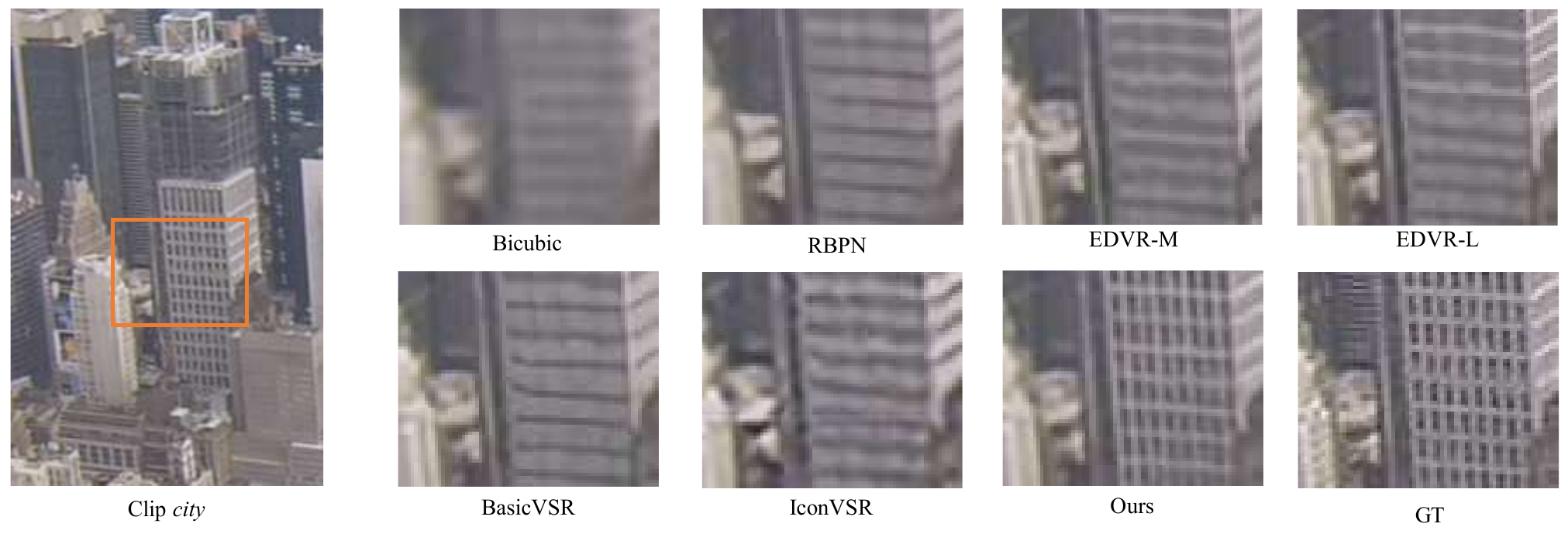}}
    \caption{ Qualitative results on different test sets. Zoom in for better visualization.}
    \label{fig6}
\end{figure*}

Vid4 dataset~\cite{vid4} has less significant motion than other test datasets and cannot fit the real video scene, which is demonstrated by the average flow magnitude (pixel/frame)\cite{haris2019recurrent}. Therefore, we test our model on Vid4 test dataset with degradation BD, to simulate a situation with motion offsets. While encountering LR frames with degradation Blur Down (BD), TCNet also achieves promising performance in terms of PSNR and SSIM compared with other state-of-the-art methods. Note that we re-implement OVSR and obtain the results that is different from the reported in their paper. The two results are both reported results in Tab.~\ref{tab:table2}. As OVSR achieves better performance than all the previous methods on Vid4 test dataset, TCNet performs best and surpass SOTA methods TTVSR and OVSR 0.03 dB and 0.04 dB respectively. It indicates that the sophistication of our method in the similar recurrent framework, and TCNet has the ability to reconstruct LR frames in whether BI or BD degradation. It is because that offsets of different degradation can be eliminated in the temporal domain by the proposed temporal consistent learning approach and the distribution of different degradation kernels can be well estimated. Moreover, to analyse the temporal consistency of restored videos, we also plot the PSNR changes over each frame on Vid4 test set. Fluctuation of PSNR result on the single frame objectively shows the robustness of the model in maintaining temporal consistency. As shown in Fig.~\ref{fig8}, PSNR curve of our method fluctuates lighter than other methods, compared with TGA (sliding-window), RRN (unidirectional recurrent network), and IconVSR (bidirectional recurrent network). Such results demonstrate the promising performance in maintaining temporal consistency of our method.

\begin{figure*}[htbp]
\begin{center}
\includegraphics[width=0.85\textwidth]{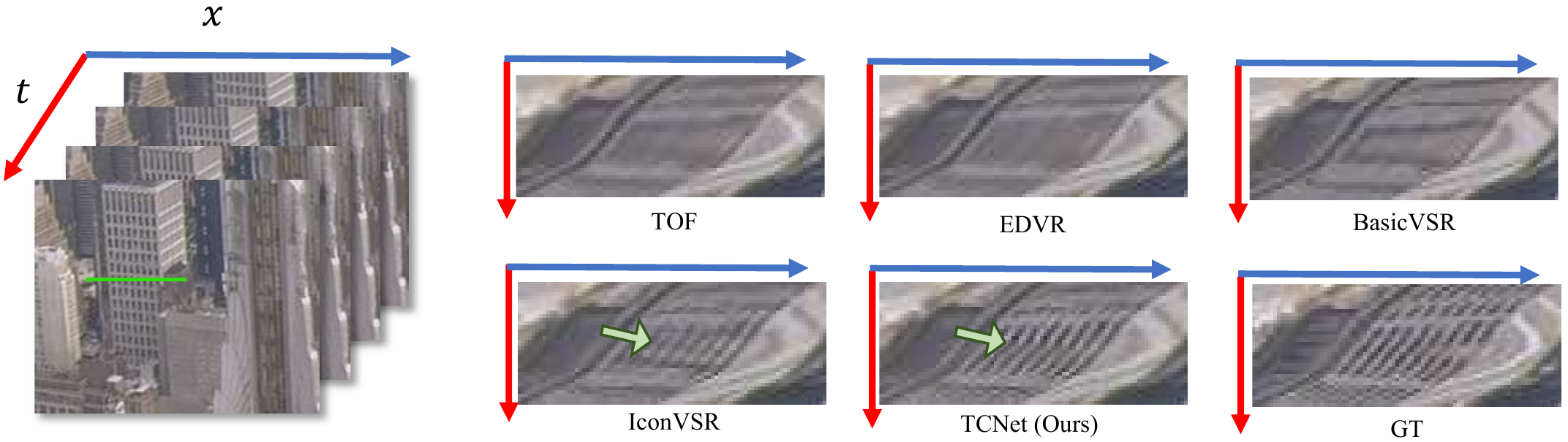}
\caption{ Comparison of temporal profile. Temporal profile is produced by recording a single-pixel line(green light) with stacking among frames. \label{fig7}}
\end{center}
\end{figure*}

We also present some visual results of reconstructed frames, which are shown in Fig.~\ref{fig6}. Compared with other methods, our network can restore more reliable and visual pleasing edge details. Besides, accurate texture details and reliable temporal information are also well maintained. Specifically, TCNet reconstructs sharper vertical edges of column cracks in Fig.~\ref{fig6}(a) and sharp details in stair edges and crevices in Fig.~\ref{fig6}(b), while other methods only produce blurry results. In Fig.~\ref{fig6}(c), TCNet produces clearer structure of building windows than other methods. This also demonstrates the effectiveness on recovering details of TCNet. Moreover, to explore the effectiveness of our method in sustaining the consistency of reconstructed results, we also examine the temporal consistency of reconstructed results subjectively. The temporal profile is generated by stacking a row of pixels vertically among coherent frames. As shown in Fig.~\ref{fig7}, TCNet produces more consistent results and fewer artifacts than other methods. Particularly, the temporal profile generated by TCNet is smooth in the junction of pixel rows, and sharp temporal textures (green arrow pointed). The consistency of the building windows over the temporal profile is too smooth and hard to discriminate in previous works. TCNet restores distinct and detailed structure of the building window in the temporal profile while other methods only produce blurry results, which indicates that the temporal consistency of restored videos can be well maintained by our method. To further validate the reconstruction quality and temporal consistency of the videos, We additionally present several restored video clips for better visualization\footnote{https://github.com/Kimsure/TCNet-for-VSR}.

\subsection{Comparisons with Short Sequence}
Some previous works\cite{TGA,wang2019edvr,yi2019progressive} just chose a part of frames as input sequence at a time, rather than took all frames into their network. It is not fair to compare TCNet with these methods regardless of the number of input frames. Thus, we retrain our method, BasicVSR and IconVSR~\cite{chan2021basicvsr} on 5 frames in REDS~\cite{nah2019ntire}, and compare with other short sequence methods, including VSRT~\cite{cao2021video}, EDVR~\cite{wang2019edvr} et al.

As shown in Tab.~\ref{tab:table3}, TCNet performs superior to other methods in both PSNR and SSIM. In REDS4 test dataset, TCNet surpasses VSRT 0.11 dB, and EDVR 0.21 dB, with the same numbers of input frames. It indicates that the proposed hybrid recurrent architecture can well utilize short-term and long-term information. Also, spatio-temporal stability module maintain the consistency in a global range whatever the number of frames. Though there is no advantage for the recurrent structure to process short sequence for information propagation, TCNet can still achieve better performance than other methods.
\begin{table*}[htbp]
\begin{center}
    \caption{Ablation study on the effectiveness of different modules and structures. Results are tested on REDS4 test set in RGB channel. \label{tab:table4}}
    \begin{tabular}{c|c|cccc|c|c}
    \hline
\begin{tabular}[c]{@{}c@{}}Recurrent\\ Architecture\end{tabular} &
  Models &
  \begin{tabular}[c]{@{}c@{}}Temporal \\
  Self-alignment\end{tabular} &
  \begin{tabular}[c]{@{}c@{}}Spatial \\
  Correlative Matching\end{tabular} &
  \begin{tabular}[c]{@{}c@{}}One-stage \\ Fusion\end{tabular} &
  \begin{tabular}[c]{@{}c@{}}Progressive \\ Fusion\end{tabular} & Params(M) &
  PSNR(dB) \\ \hline
\multirow{4}{*}{\begin{tabular}[c]{@{}c@{}}Vanilla\\ Recurrent\end{tabular}} & Model 1  &  &  & \checkmark &  & 4.7 & 30.61 \\
                                                                             & Model 2  &  & \checkmark & \checkmark &  & 5.4 & 30.67 \\
                                                                             & Model 3  & \checkmark &  & \checkmark &  & 7.1 & 30.74 \\
                                                                             & Model 4  & \checkmark & \checkmark &  & \checkmark & 7.8 & 30.92 \\ \hline
\multirow{4}{*}{\begin{tabular}[c]{@{}c@{}}Motion\\ Recurrent\end{tabular}}  & Model 5  &  &  & \checkmark &  & 6.3 & 31.42 \\
                                                                             & Model 6  &  & \checkmark & \checkmark &  & 7.0 & 31.53 \\
                                                                             & Model 7  & \checkmark &  & \checkmark &  & 8.7 & 31.59 \\
                                                                             & Model 8  & \checkmark & \checkmark &  & \checkmark & 9.4 & 31.70 \\ \hline
\multirow{4}{*}{\begin{tabular}[c]{@{}c@{}}Hybrid\\ Recurrent\end{tabular}}  & Model 9  &  &  & \checkmark &  & 6.5 & 31.48 \\
                                                                             & Model 10  &  & \checkmark & \checkmark &  & 7.2 & 31.57 \\
                                                                             & Model 11 & \checkmark &  & \checkmark &  & 8.9 & 31.68 \\
                                                                             & Model 12 & \checkmark & \checkmark &  & \checkmark & 9.6 & 31.82 \\ \hline
\end{tabular}
\end{center}
\end{table*}
\begin{table*}[htbp]
\begin{center}
    \caption{Ablation study on the influence of different fusion mechanism. Results are tested on REDS4 test set in RGB channel. \label{tab:table5}}
\begin{tabular}{c|c|c|c|c|c|c|c}
\hline
Fusion Model & $Conv + Conv$ & $Pyramid + Conv$ & $Conv + Pyramid$ & $Pyramid + Pyramid$ & Params(M) & Runtime(ms) & PSNR(dB)  \\ \hline
Model A      &     \checkmark        &                &                &     &  7.2 &  80      & 31.63 \\
Model B      &             &      \checkmark          &                &        &  8.4 &  87             & 31.69 \\
Model C      &             &                &        \checkmark        &        &  8.4 &  86             & 31.73 \\
Model D      &             &                &                &     \checkmark   &  9.6 &  94             & 31.82 \\ \hline
\end{tabular}
\end{center}
\end{table*}

\section{Ablation Study}
In this section, we conduct several ablation studies to analyse the effectiveness of different modules and structures in TCNet. All the ablation models are trained on REDS~\cite{nah2019ntire} and tested on REDS4. As shown in Tab.~\ref{tab:table4}, Model 1 to 12 denote the corresponding networks with different modules and recurrent architectures. 
\subsection{Effectiveness of hybrid recurrent architecture.}
Model 4, Model 8 and Model 12 are compared to analyse the effectiveness of the proposed hybrid recurrent architecture, where the different models denotes the corresponding recurrent structures. As shown in Tab.~\ref{tab:table4}, Model 12 achieves 0.9 dB gains than Model 4 and 0.12 dB gains than Model 8. These results indicate the effectiveness of the hybrid recurrent architecture. These improvements are derived from the fully utilization of short-term and long-term information by hybrid recurrent architecture.

\begin{figure}[htbp]
\begin{center}
\includegraphics[width=0.47\textwidth]{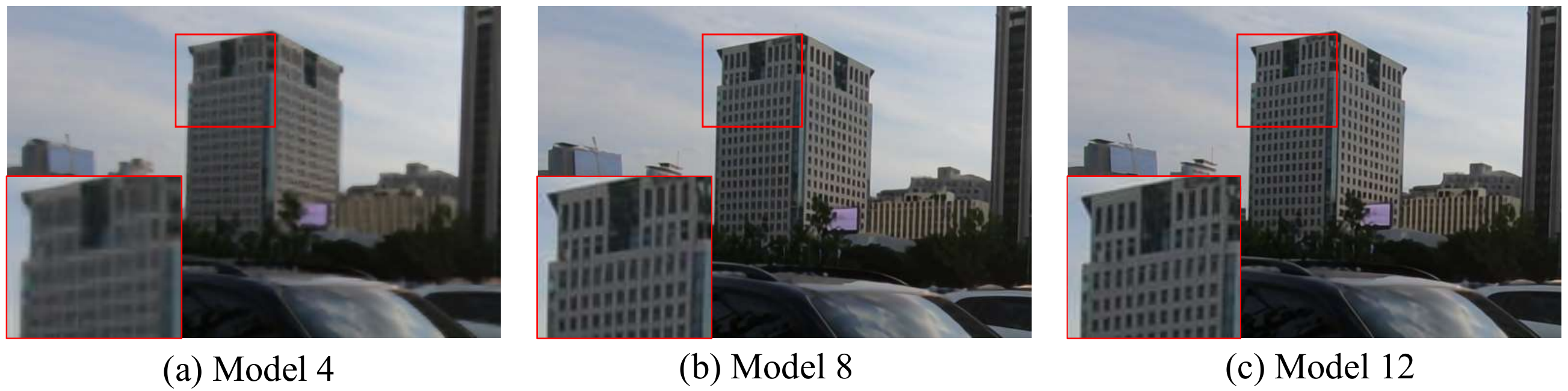}
\caption{Subjective comparison of reconstructed results with different recurrent architecture. Zoom in for better visualization.\label{fig9}}
\end{center}
\end{figure}

We also present some visual results of different recurrent architectures in Fig.~\ref{fig9}. Model 4 suffers from obvious artifacts and blurs as adjacent frames are not aligned in the information propagation and dislocated pixels overlay cause the visual parallax. In comparison, Model 8 and Model 12 both generate visually exquisite results with complete textures, sharp details, and less artifacts. Moreover, Model 12 performs better than Model 8 on restoring edges and details. The high visual quality demonstrates the importance of motion estimation. And bidirectional motion estimation can further improve the alignment accuracy. It also indicates that aligned adjacent frames by motion estimation are effective on the current frame reconstruction.

\subsection{Effectiveness of Spatio-temporal Stability Module}
In this subsection, we study the the effect of the spatio-temporal stability module. We redesign different models by removing CMB and TSB respectively. Different models and results are shown in Tab.~\ref{tab:table4}. Comparison of models w/ and w/o different blocks indicates the effectiveness of spatio-temporal stability module, no matter what the recurrent architecture is. For clarity, we take models with hybrid recurrent architecture for an example. In particular, Model 11 achieves 0.11 dB gains than Model 10, which demonstrates the effectiveness of TSB is more promotional than CMB. Model 12 achieves the best performance, with 0.14-0.34 dB gains than other models.
These results verify the effectiveness of the temporal and spatial information aggregation in the temporal domain. Besides, it should be noted that, since the operations of CMB and TSB are parallel, the absence of either lead the generated features can be only fused with the propagation features, namely one-stage fusion. Therefore, it is fair and the results are not influenced by the fusion operation. Besides, we also analyze the cost performance of each module. Model 11 surpasses Model 9 0.2 dB, with supplementary 2.4M parameters. Model 10 surpasses Model 9 0.11 dB, with supplementary 1.7M parameters. These results demonstrate that the proposed TSB and CMB can improve the reconstruction performance with affordable parameter cost.

\begin{figure}[htbp]
\begin{center}
\includegraphics[width=0.47\textwidth]{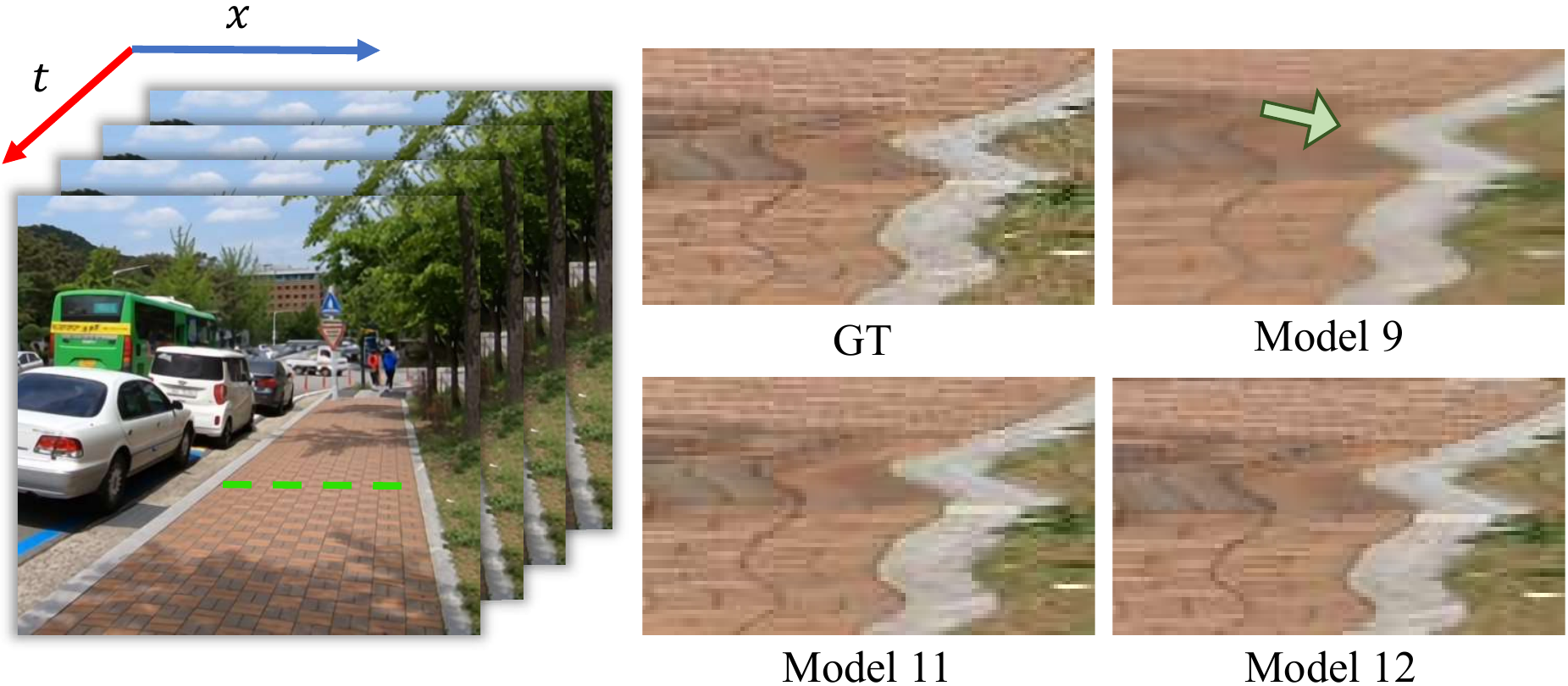}
\caption{Comparison of temporal profile w/ and w/o spatio-temporal stability module. Model 9 w/o the spatio-temporal stability module generate smooth result and the details are lost. By temporal self-aligning 3D patches across multi-frames, Model 11 and 12 both generate continuous profile. \label{fig10}}
\end{center}
\end{figure}
Moreover, visual results of temporal profile are given in Fig.~\ref{fig10} to validate the effectiveness of maintaining temporal consistency subjectively.
It is observed that Model 9 can restore the overall structure and texture information of frames. Nevertheless, the restored details in the temporal profile are too smooth, and specifically, the consistency of the white road shoulder over the temporal profile is too glossy where the green arrow points. It is because that the lack of self-alignment in the temporal domain impairs the consistency of restoration. In comparison, both Model 11 and Model 12 generate sharper details and less artifacts in the temporal profile than Model 9. It means that the temporal self-alignment can well maintain the temporal consistency.

\begin{table}[htbp]
\begin{center}
\caption{Ablation study on the pyramid levels and feature channels of fusion module. \label{tab6}}
\begin{tabular}{c|cc|c|c}
\hline
Pyramid channels and levels & C-64 & C-128 & Parameters(M) & PSNR(dB)  \\ \hline
\multirow{2}{*}{Level-2}    & \checkmark   &     & 9.29          & 31.74 \\
                            &    & \checkmark    & 9.34          & 31.75 \\ \hline
\multirow{2}{*}{Level-3}    & \checkmark   &     & 9.52          & 31.80 \\
                            &    & \checkmark    & 9.63          & 31.82 \\ \hline
\end{tabular}
\end{center}
\end{table}
\subsection{Effectiveness of Progressive Fusion}
We investigate the effectiveness of progressive fusion module in this subsection. Experimental settings and results are shown in Tab.~\ref{tab:table5}. Model A to D represent the different fusion mechanisms in the progressive fusion module. Note that $Pyramid + Conv$ is not the same as $Conv + Pyramid$. $Pyramid + Conv$ represents the first stage fusion with the pyramid structure and the second stage fusion with the convolution layers. In particular, $Conv + Pyramid$ represents the first stage fusion with the convolution layers and the second stage fusion with the pyramid structure. Other models have the same settings as before. As summarized in Tab.~\ref{tab:table5}, Model A has the worst performance, which indicates that the bad performance of fusion with simple concatenation and convolution. Compared with Model B, Model C achieves 0.04 dB gains, which demonstrates the effectiveness of the multi-stage fusion with pyramid structure between propagation features and temporally self-aligned features. Compared with Model A, Model D employs the pyramid structure to substitute a single convolution layer in both two stages, and achieves 0.19 dB gains, which indicates that progressive fusion module is effective for spatial and temporal feature fusion. The cost performance is also analyzed in the ablation study. Model C surpasses Model B 0.04 dB with equal parameters. Model D surpasses Model C 0.09 dB with only 1.2M extra parameters. These results demonstrate that the fusion performance is improved by pyramid structure with slightly larger parameters.

Furthermore, we conduct some ablation studies on the number of pyramid level and feature channel in the proposed fusion module. The results are summarized in Tab.~\ref{tab6}. When employing a pyramid fusion module at the same level, the model's performance only gains 0.01-0.02 dB with doubled channel numbers. The improvement is limited while the additional channel number is not influential to the fusion performance. Meanwhile, the increased pyramid levels get 0.07dB improvement with slightly larger parameters and the same channel numbers. It indicates that the fusion module with different scales contributes to the information aggregation and reconstruction performance.

\begin{table}[htbp]
\begin{center}
\caption{Ablation study on the pyramid levels and feature channels of fusion module. \label{tab7}}
\scalebox{0.9}{
\begin{tabular}{c|cc|c|c}
\hline
Optical flow    & \begin{tabular}[c]{@{}c@{}}w/ \\ fine-tune\end{tabular} & \begin{tabular}[c]{@{}c@{}}w/o \\ fine-tune\end{tabular} & Params(M)             &  PSNR(dB)  \\ \hline
\multirow{2}{*}{PWCNet} &   \checkmark                                                     &       & \multirow{2}{*}{17.5} & 31.77 \\
          &         &  \checkmark      &            &  30.58 \\ \hline
\multirow{2}{*}{SpyNet}   &   \checkmark       &        & \multirow{2}{*}{9.6}  &  31.82 \\   &      & \checkmark        &                       & 30.32 \\ \hline
\multirow{2}{*}{\begin{tabular}[c]{@{}c@{}}Separable\\ Flow\end{tabular}} &  \checkmark     &        & \multirow{2}{*}{14.0} & OOM     \\      &                    & \checkmark      &      & 30.62 \\ \hline
\multirow{2}{*}{RAFT}                 &   \checkmark                                                     &                      & \multirow{2}{*}{13.3} & OOM    \\               &          &  \checkmark     &          & 30.61 \\ \hline
\end{tabular}}
\end{center}
\end{table}
\subsection{Influence of Different Optical Flow Network}
In this subsection, we study the effect of different optical flow networks on the reconstruction performance. Experimental results are shown in Tab.~\ref{tab7}. In particular, we substitute the optical flow network SpyNet~\cite{ranjan2017optical} with PWCNet~\cite{sun2018PWC}, Separable Flow~\cite{zhang2021separable} and RAFT~\cite{raft}, with the same experimental settings. Note that the fine-tuned Separable Flow and RAFT are not considered while they both suffer from out-of-memory (OOM) in the training process. As summarized in Tab.~\ref{tab7}, all the optical flow networks without fine-tuning suffer from severe performance deficiency, which is up to 1.5 dB reduction compared to corresponding fine-tuned models. It is considered that the bias between VSR dataset and optical flow dataset would affect the accuracy of frame alignment. The performance of fine-tuned PWCNet is similar to SpyNet and it indicates that the bias between different datasets can be eliminated while the optical flow network is trained with the whole network. In view of the fine-tuned flow estimation module, different optical flow networks slightly affect the performance of our proposed network. It demonstrates that our TCNet is not sensitive to flow networks, while the temporal self-alignment module also contributes to the motion compensation among the video. Furthermore, it is difficult to employ the state-of-the-art methods of optical flow on VSR networks due to large parameters and memory requirements, leading to unaffordable training and deployment costs.

\section{Conclusion}
In this paper, we propose an end-to-end temporal consistency learning network for video super-resolution (TCNet). We firstly present a hybrid recurrent architecture with bidirectional motion estimation. In the information prepropagation and refinement, the presented architecture has the ability to leverage short-term and long-term information. Then, we propose a spatio-temporal stability module with two blocks, to enhance the structure stability of frames and the temporal consistency of the restored video. The spatial correlative matching block (CMB) preserves the stability within the single frame by the matching degree. The temporal self-alignment block (TSB) learns the global motion tendency to implement an adaptive warping across multiple frames, where the temporal consistency can be maintained. We further introduce a progressive fusion module to enhance the feature fusion in a multi-scale space. Comprehensive experiments on different datasets demonstrate that our TCNet performs better than other state-of-the-art methods in both long and short sequence experiments for VSR.

\bibliographystyle{IEEEtran}
\bibliography{egbib}

\end{document}